\documentclass[sigconf, nonacm]{acmart}
\AtBeginDocument{%
  }

\setcopyright{acmlicensed}
\copyrightyear{2018}
\acmYear{2018}
\acmDOI{XXXXXXX.XXXXXXX}
\acmConference[Conference acronym 'XX]{Make sure to enter the correct
  conference title from your rights confirmation email}{June 03--05,
  2018}{Woodstock, NY}
\acmISBN{978-1-4503-XXXX-X/2018/06}

\acmSubmissionID{180}

\usepackage{float}
\usepackage{xspace} % 自动处理宏命令后的空格
\usepackage{multirow}
\usepackage{enumitem}
\usepackage{makecell}
\usepackage{multicol}
\usepackage[capitalize]{cleveref}
\crefname{section}{Sec.}{Secs.}
\Crefname{section}{Section}{Sections}
\Crefname{table}{Table}{Tables}
\crefname{table}{Tab.}{Tabs.}

\newcommand{\myparagraph}[1]{{\vspace{0.2\baselineskip} \noindent \bfseries #1}}
\newcommand{\eg}{\textit{e.g.}\xspace}     % 举例 (for example)
\newcommand{\ie}{\textit{i.e.}\xspace}     % 即 (that is)
\newcommand{\etc}{\textit{etc.}\xspace}    % 等等 (and so forth)
\newcommand{\vs}{\textit{vs.}\xspace}      % 对决/与...相比 (versus)
\begin{document}

\title{StructGen: Disambiguating Multi-Reference Image Generation via Structured Context Modeling}

\newcommand{\cvprauthorblock}{%
\parbox{0.98\textwidth}{%
\centering

{\fontsize{12pt}{15pt}\selectfont
Jianing Peng\textsuperscript{1,3,5*}\quad
Mengyu Wang\textsuperscript{1,3*}\quad
Henghui Ding\textsuperscript{2\textdagger}\quad
Zixiang Li\textsuperscript{1,3}\quad
Ting Liu\textsuperscript{5}
}\\[0.45em]

{\fontsize{12pt}{15pt}\selectfont
Xiaochao Qu\textsuperscript{5}\quad
Luoqi Liu\textsuperscript{5}\quad
Yao Zhao\textsuperscript{1,3}\quad
Yunchao Wei\textsuperscript{1,3,4\textdagger}
}\\[0.65em]

{\fontsize{12pt}{15pt}\selectfont
\textsuperscript{1}Institute of Information Science,
Beijing Jiaotong University, Beijing, China\\
\textsuperscript{2}Institute of Big Data,
Fudan University, Shanghai, China \\
\textsuperscript{3}Visual Intelligence + X International Joint Laboratory of the Ministry of Education, Beijing, China\\
\textsuperscript{4}Beijing Academy of Artificial Intelligence, Beijing, China \\
\textsuperscript{5}MT Lab, Meitu Inc, Beijing, China
}\\[0.45em]

{\fontsize{12pt}{15pt}\selectfont
\textsuperscript{*}Both authors contributed equally to this research.
\qquad
\textsuperscript{\textdagger}Corresponding authors.
}

}}

\author{\texorpdfstring{\cvprauthorblock}
{Jianing Peng, Mengyu Wang, Henghui Ding, Zixiang Li, Ting Liu,
Xiaochao Qu, Luoqi Liu, Yao Zhao, and Yunchao Wei}}

\affiliation[obeypunctuation=true]{%
  \institution{\mbox{}}
  \city{\mbox{}}
  \country{\mbox{}}
}

% Prevent acmart from printing the custom author block again.
\authorsaddresses{}

\renewcommand{\shortauthors}{Jianing Peng, et al.}

\begin{abstract}

Multi-reference image generation aims to synthesize images by integrating attributes from multiple reference images under textual instructions.
As the number of references increases, the task necessitates complex semantic comprehension, such as correctly associating attributes with the intended subjects and planing out coherent spatial arrangement between subjects and their environments.
Existing approaches, which rely solely on natural language instruction, often fail to capture these complex intentions precisely, leading to semantic misalignment and inconsistent generation.
We identify two key factors behind these limitations: natural language instructions are often verbose and ambiguous, and high-quality multi-reference data is scarce.
To address these issues, we propose StructGen, which employs a structured, dictionary-like format to encode multiple reference images, thereby enabling explicit and unambiguous specification of generation intentions.
To support this design, we construct a structured dataset based on high-quality real images and develop a corresponding training framework, along with a dedicated benchmark for challenging multi-reference scenarios.
Extensive experiments on both public benchmarks and our proposed benchmark demonstrate that StructGen consistently outperforms existing methods on both semantic alignment and detailed reference-generation consistency, especially under complex instructions with multiple references.
The code is available at \href{https://jianingpeng0382.github.io/StructGen/}{https://jianingpeng0382.github.io/StructGen/}.

\end{abstract}



%   \includegraphics[width=\textwidth]{figs/intro_layout_1.png}
%   \caption{Failure case for complex semantic relationships across reference images and scene and character consistency.}
\begin{teaserfigure}
    % \vspace*{-1.5em}
    \centering
    \includegraphics[width=\linewidth]{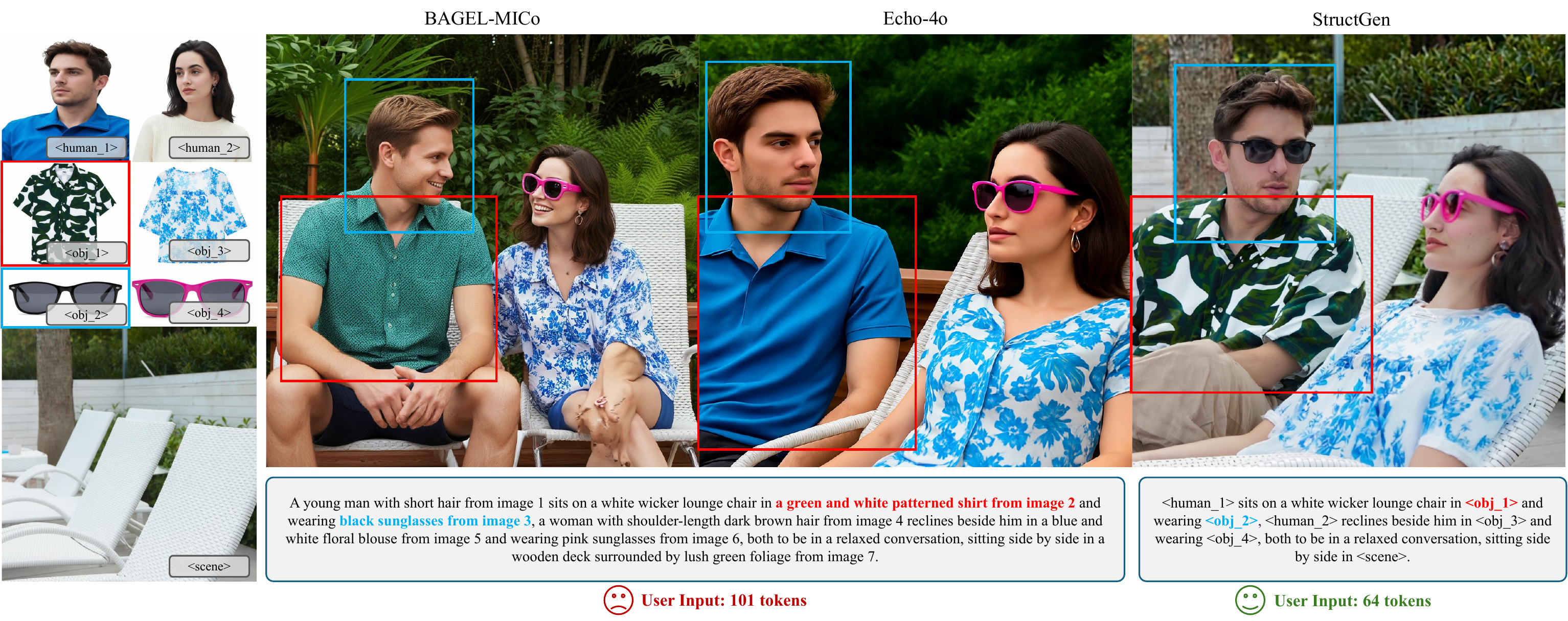}
    % \vspace{-1em}
    % \caption{Qualitative comparison of existing methods and our StructGen. Existing methods suffer from (1)~verbose and ambiguous free-form prompts (101 tokens), (2)~incorrect attribute--subject associations (clothing and glasses are neglected), and (3)~degraded consistency with reference (facial identity and background). StructGen encodes references in a dictionary-like manner with a compact structured prompt (64 tokens) and is trained on structured data curated from real-world images, correctly preserving both identity and attribute--subject associations.}
    \caption{Qualitative comparison on a challenging example with multiple references. Through structured context modeling, StructGen enables concise and unambiguous intent specification (only 64 tokens), accurate attribute–subject association (correctly generating the clothing and glasses), and strong consistency with references (facial identity and background scene).}
    \label{fig:teaser}
\end{teaserfigure}

\maketitle

\section{Introduction}
\label{sec:intro}

% \textbf{Multi-reference image generation and its compositional challenges}

% \textbf{Limitations of free-form prompting in multi-reference settings}

% \textbf{From prompt ambiguity to structured context representation}

% \textbf{Contributions}

Multi-reference image generation aims to synthesize an image by incorporating visual content from multiple reference images under the guidance of a textual instruction.
As a pivotal area of research, human-centric scenarios have garnered substantial attention, which enables users to flexibly compose specific human, objects, and backgrounds into a coherent scene.
This task holds significant potential for a wide range of real-world applications, such as e-commerce display, virtual try-on, and storyboard creation.

Recently, with the rapid scaling of data and model capacity, generative models~\cite{wu2025omnigen2, gpt4o, wu2025less, xia2025dreamomni2, cui2025emu35, chen2025blip3o, li2025visualcloze, chen2025multiref} have begun to exhibit preliminary capabilities in multi-reference image generation.
Despite these advances, existing models still face notable limitations with multiple references, struggling to accurately follow user instructions and faithfully reflect the visual details of the references.
On the one hand, as the number of reference images increases or as instructions involve more complex associations among them, models struggle to correctly interpret and ground the intended semantics, often leading to incorrect associations between attributes and subjects. As illustated in~\cref{fig:teaser}, when presented with multiple human and their associated attire, existing models~\cite{wei2025mico,ye2025echo} fail to represent the desired clothing and glasses of one individual in the generation results.
On the other hand, current models frequently fail to preserve fine-grained details, resulting in inconsistencies in appearance, such as degraded facial identity or variations in the background environment, as shown in~\cref{fig:teaser}.
% , existing methods suffer from weak facial identity consistency and noticeable shifts in the background environment.

In this work, we argue that these limitations primarily stem from two underlying factors.
First, the current task formulation for multi-reference image generation is inadequate. Existing methods~\cite{wu2025less, tan2025ominicontrol, xia2025dreamomni2, wu2025omnigen2} \textbf{rely solely on natural language instructions} to specify users' intention.
As the number of references increases, to describe the complex inter-reference relationships, such instructions often become verbose and ambiguous (as shown in~\cref{fig:teaser}), posing significant challenges for both user-side description and model-side understanding.
Confronted with this issue, we seek to explore a more efficient and accurate context format that outperforms plain natural language.
Therefore, we propose StructGen, a multi-reference image generation framework with \textbf{structured context modeling mechanism}.
StructGen models multiple reference images in a structured, dictionary-like manner, where each image is assigned a unique identifier.
Leveraging this dictionary, we propose identifier-based instruction, which directly denotes references by referencing their corresponding identifiers to specify how they should be composed.
This structured format mirrors the human intuition of pointing at a reference while giving instructions, and the identifiers serve as a metaphorical pointer, enabling explicit and unambiguous specification of desired generation targets.
% This structured format enables explicit and unambiguous specification of each reference and their interactions, thereby improving semantic alignment during generation.

% making it difficult for models to accurately interpret and ground attributes and objects across references. Furthermore, these lengthy and intricate descriptions also impose a cognitive burden on users, limiting the practicality of such systems.
% which may consist of complex inter-image relationships for multiple references. As relational complexity increases, such instructions often become verbose and ambiguous (as shown in~\cref{fig:teaser}), making it difficult for models to accurately interpret and ground attributes and objects across references. Furthermore, these lengthy and intricate descriptions also impose a cognitive burden on users, limiting the practicality of such systems.
% To address this issue, we propose StructGen, a multi-reference image generation framework with \textbf{structured context modeling mechanism}.

% Specifically, we assign a unique identifier to each reference image, and interleave these identifiers with corresponding reference image tokens, encoding them in a dictionary-like manner. The textual instruction then refers to these identifiers to explicitly specify how these references should be composed, rather than relying purely on natural language instruction. This structured representation enables explicit and unambiguous specification of each reference and their interactions, thereby improving semantic grounding and reducing ambiguity during generation.

\begin{figure}[t]
    \centering
    \includegraphics[width=0.98\linewidth]{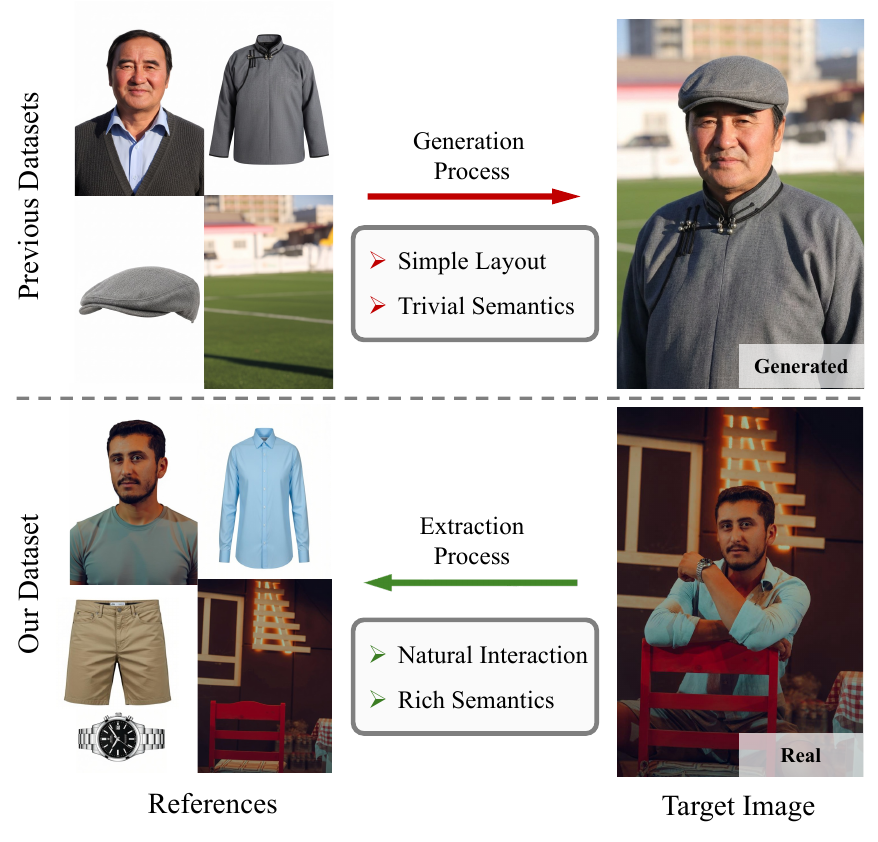}
    \vspace{-1em}
    \caption{Compared with previous datasets, ours preserves rich semantics from real-world images, including flexible human postures and human-environment interaction.}
    \vspace{-1em}
    \label{fig:intro_2}
\end{figure}

Another key limitation lies in the \textbf{lack of high-quality data} for multi-reference generation. In real-world scenarios, paired image sets suitable for this task is extremely scarce and difficult to collect.
% Early datasets primarily depend on text-to-image models to generate paired data, which are largely restricted to single-reference settings~\cite{brooks2023instructpix2pix, jiang2025anyedit, zhao2024ultraedit, hui2024hq}. 
Recent approaches~\cite{ye2025echo, wei2025mico} attempt to construct multi-reference training data using state-of-the-art closed-source generative models (\eg, GPT-4o~\cite{gpt4o} and NanoBanana~\cite{google2025nanobanana}).
They first collect individual objects as references, and then compose them into a single image through generation.
However, such distillation-manner pipeline is inherently constrained by the model biases. As illustrated in~\cref{fig:intro_2}, the resulting images often exhibit close-up views with dominant foreground subjects and minimal background context, while meaningful interactions between subjects and their environments are limited.
Consequently, these datasets struggle to facilitate the model learning of diverse spatial arrangements and complex interactions.
To overcome this limitation, we introduce a \textbf{structured data curation pipeline} based on real-world images. Unlike prior approaches that directly use generated images as training target, our method instead takes real images as targets and parses each image into multiple structured fields of references entities.
This pipeline preserves the inherent semantic richness and diversity of real-world scenes (\eg, the man's interaction with the chair in~\cref{fig:intro_2}).
The structured design is not only aligned with our framework, but also alleviates the difficulty of extracting individual reference entities, enabling more consistent and reliable parsing of visual elements.
As a result, it leads to higher fidelity and improved consistency between reference images and training targets.% To overcome this limitation, we introduce a \textbf{structured data curation pipeline} based on real-world images. Unlike prior distillation-based approaches that directly use generated images as supervision, our method instead takes real images as targets and parses each image into multiple structured fields through a carefully designed pipeline, from which corresponding entities are extracted as reference images.
% This structured design aligns naturally with our overall framework and enables consistent and reliable reference construction, maintaining strong consistency between references and targets.
% As a result, models trained on such data have a higher upper bound in generation consistency and the ablility to capture fine-grained details and interactions.

Furthermore, we construct a new benchmark to evaluate StructGen and existing models. Compared to existing benchmarks~\cite{ruiz2023dreambooth, pengdreambench++, wu2025omnigen2, chen2025xverse, wei2025mico}, our work targets more challenging scenarios in human-centric multi-reference image generation.
It incorporates diverse references, including human, attire, objects, and scenes, enabling a more comprehensive evaluation of model capabilities.
In addition, our benchmark designs metrics that separately assess the capability of prompt following, subject consistency, and facial ID consistency, providing a more fine-grained analysis.

Our contributions are summarized as follows:
\begin{itemize}[leftmargin=*]
    \item We propose StructGen, a framework for multi-reference image generation that formulates the reference images and instructions as structured context, enabling, explicit, unambiguous, and convenient specification of generation intention.
    \item We develop a high-quality data construction pipeline, featuring structured annotations, strong reference–target consistency, and real-world semantic richness, along with a dedicated benchmark towards challenging human-centric scenarios.
    \item Experiments show that StructGen outperforms existing methods on both public benchmark and our proposed benchmarks, demonstrating clear advantages in semantic alignment as well as generation consistency across multiple dimensions.
\end{itemize}

\section{Related Work}
\label{sec:related}

\subsection{Reference-based Image Generation}
% In recent years, diffusion models have become a central paradigm for modern generative tasks, especially in image synthesis. As the foundation of modern text-to-image models, they have substantially improved visual quality, prompt alignment, and controllability, making high-fidelity image generation increasingly practical. Beyond text-to-image generation, diffusion-based models have also supported a broad range of conditional generation settings, including image editing, visual instruction following, and multimodal generation.  These advances provide a strong generative backbone for increasingly complex generation tasks. However, as conditioning signals become richer and more diverse, the challenge is no longer only how to synthesize high-quality images, but also how to represent and organize the input conditions in a way that supports precise generation. This issue is particularly important in multi-reference image generation, where models must integrate multiple visual sources under compositional constraints.
Reference-based image generation aims to synthesize images by preserving visual content from reference images while satisfying a target instruction. Early studies mainly focus on single-reference settings, such as subject-driven generation~\cite{ruiz2023dreambooth, ye2023ipadapter, li2023blipdiffusion} and image editing~\cite{zhang2023magicbrush, he2025freeedit, zhao2024ultraedit, jiang2025anyedit, pengdreambench++, isola2017pix2pix, gal2023textualinversion, brooks2023instructpix2pix, hui2024hq, wang2024region, li2025dci, hu2025dcedit}. 
% Benefiting from the rapid progress of diffusion models and controllable generation techniques~\cite{zhang2023controlnet}, these methods have substantially improved generation quality and reference fidelity. 
More recently, researchers have extended this setting to multi-reference image generation~\cite{tan2025ominicontrol, wu2025less, xiao2025omnigen, xia2025dreamomni2, wu2025omnigen2, chen2025xverse}, where models integrate contents from multiple images into one. 
% Compared with single-reference generation, this task is substantially more challenging, since the model must not only preserve the content of each reference, but also correctly handle the relationship across different images and plan out reasonable and aesthetic spatial layouts. 
Existing studies have explored multi-image composition through model adaptation~\cite{wang2024msdiffusion, xiao2025omnigen} and training-free methods~\cite{zhou2024storydiffusion}. These studies show that recent generative models have begun to exhibit preliminary capabilities in multi-reference generation. However, existing approaches still struggle to accurately follow generation instructions and preserve fine-grained details from references, leading to incorrect attribute-subject associations and degraded appearance consistency.

\subsection{Reference-based Dataset and Benchmark}
The progress of reference-based generation is closely tied to the availability of paired training data and reliable evaluation protocols. 
Early methods~\cite{kumari2023customdiffusion} construct multi-concept paired data through parameter-efficient fine-tuning on limited reference images. More recent datasets increasingly rely on strong generative models to synthesize reference-target pairs at scale, either through in-context generation~\cite{tan2025ominicontrol, wu2025less, xia2025dreamomni2} or powerful proprietary models~\cite{ye2025echo, wei2025mico}, such as GPT-4o~\cite{gpt4o} and NanoBanana~\cite{google2025nanobanana}.
Although these datasets substantially produce multi-reference and target pair, they inevitably inherit biases from upstream models, failing to capture the diverse spatial arrangements and complex subject-environment interactions inherent in real-world scenes.
On the evaluation side, benchmarks have evolved from subject-driven settings~\cite{ruiz2023dreambooth, pengdreambench++} to compositional~\cite{li2024genaibench, huang2023t2icompbenc} and multi-reference evaluation~\cite{wu2025omnigen2, chen2025xverse, wei2025mico}. However, for human-centric scenarios, current benchmarks still lack specialized designs in reference configurations and metrics, making it difficult to comprehensively assess model performance.

\subsection{Unified Multimodal Model}
Recently, unified multimodal models that combine visual understanding and image generation within a shared framework have been actively explored~\cite{team2024chameleon, sun2024emu3, xie2024showo, xie2025showo2, wu2025janus, chen2025januspro, ma2025janusflow, zhou2024transfusion, deng2025emerging, chen2025blip3o, cui2025emu35, he2025plangen, wu2025omnigen2, wu2025qwen, gpt4o}. Existing approaches mainly unify the two capabilities through autoregressive modeling over discrete visual tokens~\cite{team2024chameleon, sun2024emu3, cui2025emu35}, diffusion-based generation within the language model~\cite{xie2024showo, xie2025showo2}, or hybrid architectures that decouple understanding and generation pathways while sharing a common backbone~\cite{zhou2024transfusion, wu2025janus, chen2025januspro, ma2025janusflow, deng2025emerging}. Several approaches further employ frozen LLMs for understanding with an additional diffusion transformer~\cite{peebles2023dit} for generation~\cite{wu2025omnigen2, pan2025metaquery, chen2025blip3o}. These unified models provide a natural backbone for multi-reference generation, and recent work has shown emerging in-context generation capabilities where reference images alongside instructions guide compositional synthesis~\cite{wu2025less, wu2025omnigen2}.
Among them, BAGEL~\cite{deng2025emerging} introduces a Mixture-of-Transformers architecture with flow-matching based generation, which supports interleaved image-text input and is well suited to our structured input format.
\begin{figure*}[ht]
    \centering
    \includegraphics[width=\textwidth]{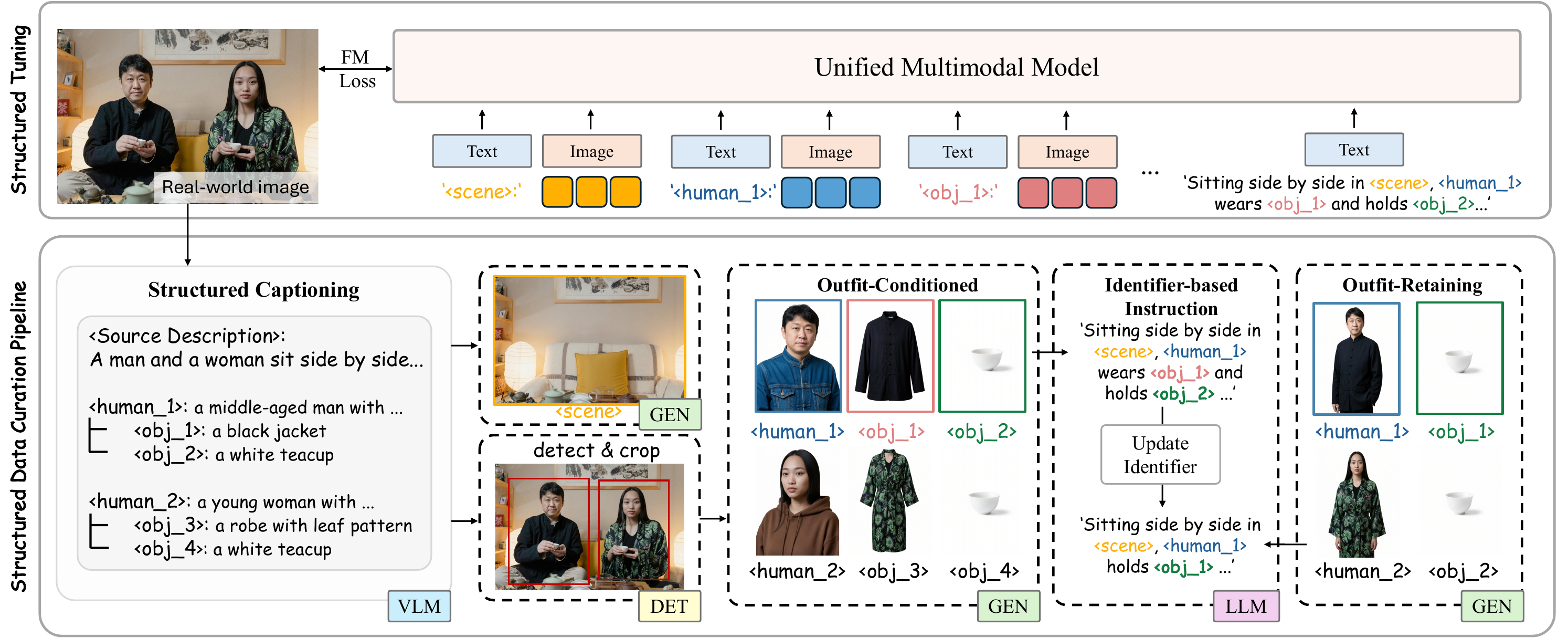}
    \vspace{-2em}
    \caption{StructGen consists of Structured Tuning (top) and Structured Data Curation (bottom). The data curation pipeline automatically transforms real-world images into structured annotated training data, with each image referenced through an identifier (\eg, \(\langle human\_1 \rangle\)). The tuning process then facilitates the model adapting to this structured context format.}
    % \vspace{-1em}
    \label{fig:method}
\end{figure*}

\section{StructGen}
\label{sec:method}
% In this section, we present \textbf{StructGen} as an integrated framework with four components. We first introduce \textbf{Structured Context Modeling}, which reformulates the task by decomposing the input into a structured reference context and a generation instruction. We then present the \textbf{StructGen Dataset}, a human-centric multi-reference dataset constructed from real images with structured annotations for references, relations, and scene context. Based on this representation, we develop \textbf{Structured Instruction Tuning}, which trains the model with placeholder-based grounding and structured context dictionaries for more precise reference binding and composition. Finally, we establish the \textbf{StructGen Benchmark}, a benchmark designed to evaluate generation precision in challenging multi-reference settings from the aspects of reference preservation, compositional correctness, and overall instruction following.

\subsection{Structured Context Modeling}
\label{sec:3_1}
\myparagraph{Preliminary}
We begin with a brief review of BAGEL, upon which our StructGen is built.
BAGEL process reference images and text instructions by concatenating their encoded tokens into a single context sequence. Given a set of reference images \(\{I_{\mathrm{ref}}^i\}_{i=1}^{N}\), each reference image is first encoded by an image tokenizer \(\mathcal{E}(\cdot)\):
\begin{equation}
\mathbf{z}_{\mathrm{ref}}^i = \mathcal{E}(I_{\mathrm{ref}}^i), \quad i = 1, \ldots, N, 
\end{equation}
where \(\mathbf{z}_{\mathrm{ref}}^i\) denotes the encoded image tokens and are subsequently flattened into a 1D sequence.
Similarly, the text instruction \(c\) is processed by a text tokenizer. Tokens from both modalities are then concatenated along the sequence dimension in a fixed order to jointly form the model’s complete context \(\mathbf{z}_c\):
\begin{equation}
\mathbf{z}_c = \mathrm{Concatenate}(\mathbf{z}_{\mathrm{ref}}^1, \mathbf{z}_{\mathrm{ref}}^2, \ldots, \mathbf{z}_{\mathrm{ref}}^N,c ). \label{equ:concat_context}
\end{equation}

During generation, BAGEL treats the concatenated context as a condition, extracting information from it via attention~\cite{vaswani2017attention} to generate the target image. 
Notably, although the architecture of BAGEL supports interleaved input format, both its pretrained weights and subsequent fine-tuning works, such as Echo-4o~\cite{ye2025echo} and BAGEL-MICo~\cite{wei2025mico}, have not exploited this capability. 
We argue that this naive input format acts as a bottleneck for multi-reference generation.
The fixed concatenation order, images followed by text, flattens the independence between input elements.
Consequently, instructions must employ semantic cues to denote a specific reference, such as reference indexing and detailed descriptions (e.g., ``\textit{the green and white patterned shirt from image 2}'' in~\cref{fig:teaser}). 
However, such information is already implicitly contained within the reference images.
Requiring it to be redundantly restated not only imposes unnecessary burden on users but also complicates model understanding.
This motivates our following exploration of a more efficient context modeling approach as alternative.

\myparagraph{Structured Context}
% To address the difficulty of clearly and efficiently specifying generation targets, we propose a simple yet effective reformulation of the input context. 
Instead of directly concatenating reference image tokens and text tokens in a fixed order, we redefine the context \(\mathbf{z}_c\) as a structured representation composed of two components: a \textbf{reference dictionary} and an \textbf{identifier-based instruction}. 
The former presents the available visual elements, while the latter specifies how these elements should be composed in the target image.
Specifically, in the reference dictionary, each reference image \(I_{\mathrm{ref}}^i\) is assigned a unique identifier \(id_i\), formulated as:
\begin{equation}
\mathcal{D}_{\mathrm{ref}} = \{\, id_i : \mathbf{z}_{\mathrm{ref}}^i \,\}_{i=1}^{N},
\end{equation}
where \(id_i\) acts as a key and \(\mathbf{z}_{\mathrm{ref}}^i\) as its corresponding value. 
On top of this dictionary, we introduce an identifier-based instruction \(c_{\mathrm{id}}\), which explicitly refers to these identifiers to specify the desired generation results. 
The overall structured context is defined as:
\begin{equation}
\mathbf{z}_c^{\mathrm{struct}} = \mathrm{Concatenate}(\mathcal{D}_{\mathrm{ref}},\, c_{\mathrm{id}}).
\end{equation}

Additionally, as we primarily focus on human-centric scenarios in this work, we instantiate the identifiers as \(\langle \mathrm{human}_i \rangle\), \(\langle \mathrm{obj}_i \rangle\), \etc
More details can be found in~\cref{fig:method} and ~\cref{sec:3_2}.
By introducing identifiers as explicit anchors, our approach avoids entangling content and relations in ambiguous natural language, transforming cross-reference association from implicit understanding and reasoning into an explicitly searching, thus leading to more accurate reference grounding and more consistent generation.

\subsection{Structured Data Curation Pipeline}
\label{sec:3_2}

To support the training of model with structured context modeling, it is necessary to first construct a training dataset with such annotations.
Considering the deficiencies in semantic diversity among current open-source datasets (see~\cref{fig:intro_2}), we moved beyond merely adapting existing datasets. Instead, we constructed a novel data curation pipeline using real-world images to produce structured annotations and highly consistent references.
\cref{fig:method} shows the entire pipeline, which consists of the 4 stages: (1) source image collection, (2) reference-aware captioning, (3) reference extraction, and (4) identifier-based instruction synthesis.
In the following, we elaborate on each stage of the pipeline.

\myparagraph{Source Image Collection}
Towards high-quality real-world imagery with diverse semantics, we curated an initial data pool by collecting royalty-free assets from the Internet.
To ensure the quality and diversity of the training data, we employ a series of filtering strategies on the initial data pool.
First, we remove duplicate images based on perceptual hashing, which effectively eliminates redundant samples and improves the overall diversity.
Second, as our work focuses on human-centric scenarios, we apply a face detection model, InsightFace~\cite{deng2018arcface}, to filter our images without detectable faces, or those containing faces with low confidence scores or extremely small sizes. This step ensures that retained samples contain clearly visible human subjects.
Finally, we further refine the data pool by assessing face quality. While the previous step guarantees the presence of faces, it does not account for issues such as blur or large-portion occlusion. Therefore, we employ Q-Align~\cite{wu2024q} to evaluate the quality of detected face regions and filter out samples with low scores.
Since multi-human data is under-represented in the initial data pool, we further collect more data especially for multi-human scenarios.
We employ a YOLO-based model~\cite{Jocher_Ultralytics_YOLO_2023} to detect human subjects and InsightFace to detect faces.
Images are retained only when the detected human count is greater than one and matches the face count.
Through these filtering steps, we obtain a cleaner and more reliable data pool with high-quality human-centric visual content, which consists of 25,718 samples, serving as a foundation for subsequent pipeline.
% , representing approximately 6.8\% of the total data initially downloaded.
% This data pool provides a strong foundation for subsequent structured data construction.
% To construct training data under the structured context formulation, we design a curation pipeline consists of three stages: ID-grounded captioning, reference image construction, and structured text construction. Their outputs are merged into unified training samples with synchronized visual and textual supervision.

\myparagraph{Reference-Aware Captioning}
In this stage, for each source image, we identify three kinds of entities as potential references: human, belongings, and scene.
In contrast to existing datasets, we avoid defining a generic ``object'' category. Instead, we focus on human belongings, such as clothing and handheld items. This approach can facilitate the model's learning of semantic connections and interactions between humans and objects, rather than simple independent combinations, which better aligns with the task requirements of human-centric scenarios.
The scene represents the clean background environment excluding the human and belongings.
We apply a VLM (Qwen3-VL~\cite{Qwen3-VL}) to produce per-human, per-belonging, and scene descriptions in a JSON format.
Parallel to these fields of each entity, the VLM also yields a full-image caption, capturing its comprehensive global information and associations across entities, which remains in natural language format at this stage.
We then assign each entity a unique identifier, such as \(\langle \mathrm{human}_i \rangle\), \(\langle \mathrm{obj}_i \rangle\), and \(\langle \mathrm{scene} \rangle\), as shown in~\cref{fig:method}.

\myparagraph{Reference Extraction}
To conduct references from the collected source images, we employ a generative model (Qwen-Image-Edit~\cite{wu2025qwen}) to perform entity extraction and design tailored extraction prompts for each category of references based on the descriptions produced in the previous stage.
Specifically, for fine-grained appearance and identity preserving during extraction, our extraction prompts adhere to two key principles. \textbf{Background isolation}: placing entities on neutral backgrounds to eliminate environmental interference with their shape and identity. \textbf{Pose maintaining and perturbation}: preserving the general pose of subjects but mandate a minor and random shift, to prevent the model from learning a copy-paste bias that could arise from perfectly identical poses.

% Considering the real-world requirements in human-centric scenarios, we designed two distinct tasks with different extraction strategy: Outfit-Conditioned and Outfit-Retaining.
% In the former, when extracting human subject, we randomly sample a outfit description from a predefined bank and prompt the generative model to produce an outfit-swapped instance of the original human as reference.
% Meanwhile, both the original outfit and handheld objects are extracted as separate references.
% In the latter, human is extracted while keeping the original outfit intact.
% Outfits are no longer extracted separately as they are already part of the human reference, while handheld objects still serve as references.
% In both tasks, the scene background is always extracted as a reference. 
Considering real-world requirements in human-centric scenarios, we design two complementary tasks with different inputs: \textbf{Outfit-Conditioned} and \textbf{Outfit-Retaining}.
In the former, when extracting human subjects, we randomly sample an outfit description from a predefined bank and prompt the generative model to produce an outfit-swapped version of the original human as the reference.
Meanwhile, both the original outfit and handheld objects are extracted as separate references.
In the latter, the human subject is extracted while preserving the original outfit.
As the clothing is already part of the human reference, it is no longer treated as an independent reference, whereas handheld objects remain as separate references.
In both settings, the scene background is consistently extracted as an additional reference.
~\cref{fig:method} illustrates an example for both tasks.
Furthermore, for multi-human samples, to avoid mismatching during extraction, we first crop the per-human regions in the images using the bounding box detected in the first stage.
The extraction is then performed on the croped view, which significantly reduces cross-human interference compared to extracting from the original full image.
For further details regarding the extraction prompts for each task and reference category, please kindly refer to our supplementary material.
% We collect candidate real images containing two or three persons from Pexels and apply a two-stage filtering pipeline. First, YOLO is used for person detection and InsightFace for face detection; images are retained only when the detected person count matches the target, the total person bounding-box area exceeds a minimum ratio of the image, all faces are visible, and the maximum face yaw angle stays within a threshold. Second, we align each structured human ID with its corresponding detected person region via VLM-based matching: the VLM is prompted with numbered bounding boxes on the original image to associate each \(\langle \mathrm{human}_i \rangle\) with a specific person. Based on this alignment, we generate per-person crop images and use them as localized inputs for the edit-based extraction model, which significantly reduces cross-person interference compared to extracting from the full image. The remaining annotation steps reuse the same single-person pipeline described above.

\begin{figure}[t]
    \centering
    \includegraphics[width=0.98\linewidth]{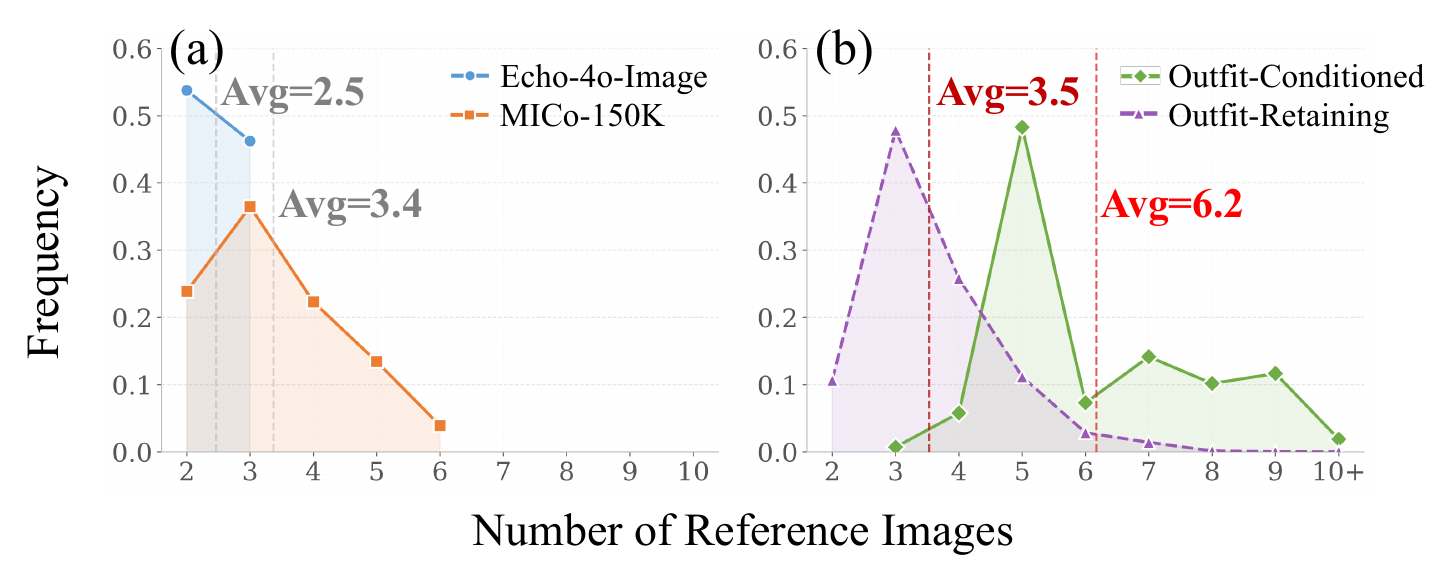}
    \vspace{-1em}
    \caption{Statistics on the number of reference images per sample across previous (a) and our (b) datasets.}
    \vspace{-1em}
    \label{fig:dataset_compare}
\end{figure}

Upon completing the extraction process, we further perform post-filtering to ensure the quality of the final dataset. While the consistency of belongs and scene elements in the extracted references is generally high, some human references still exhibit identity inconsistencies.
To address this issue, we employ InsightFace to extract facial identity embeddings and apply the Hungarian algorithm~\cite{kuhn1955hungarian, munkres1957algorithms} to establish a strict one-to-one correspondence between extracted references and their original images.
A sample is retained only if a valid one-to-one matching can be achieved and all matched face pairs exhibit high similarity scores, ensuring reliable identity alignment across references and targets.
After this stage, we obtain a final dataset consisting of 15,965 high-quality samples, with the distribution of references count shown in~\cref{fig:dataset_compare}.
% Upon completing the extraction process, we further perform post-filtering to ensure the quality of the final dataset. While the consistency of belongs and scene elements in the extracted references is generally high, some human references exhibit identity inconsistencies.
% To address this issue, we employ InsightFace for facial ID embedding extraction and apply the Hungarian algorithm to enforce a strict one-to-one correspondence between extracted references and their original images. Only samples with high facial ID similarity scores are retained.
% After this stage, we obtain a final dataset consisting of 15,965 high-quality samples, with the distribution summarized in~\cref{tab:dataset_overview}.

% \input{tab/dataset_review}

\myparagraph{Identifier-based Instruction Synthesis}
In this stage, we employ an LLM to synthesize identifier-based instructions.
For the Outfit-Conditioned task, the LLM takes as input the structured descriptions produced in the second stage, including both entity-level descriptions for each reference and the full-image caption.
It then replaces the specific entities mentioned in the full-image caption with their corresponding identifiers, transforming the full-image caption into the desired identifier-based instruction.
For the Outfit-Retaining task, the instruction synthesis process follows a similar procedure, with an additional requirement that the model recognizes the clothing items and excludes them from the references, as clothing is no longer treated as independent reference.
In addition, we prompt the LLM to rewrite independent detailed captions for each reference item, ensuring that the information in the caption across different references does not overlap. These captions are also utilized in the subsequent training process, as detailed in~\cref{sec:3_3}.

These stages collectively contribute to produce all the essential components of every training sample, including the target image, reference images, identifier and caption of each reference, and the identifier-based instruction. 
Compared to existing datasets, data produced by this pipeline maintains high-fidelity real-world semantics and structured annotations. Furthermore, it also exhibits a greater mean value and a wider distribution spectrum of the reference count per sample, as depicted in~\cref{fig:dataset_compare}.

% We then generate fine-grained appearance descriptions for each entity. For each human, Qwen-VL produces both a part-level description covering individual components (face, clothing, accessories) and a whole-level description capturing the overall appearance. For belongings, wearable items (e.g., clothing) are distinguished from non-wearable objects (e.g., carried items) via keyword-based rules. Based on this classification, we use Qwen3 to rewrite the part-human ID-based instruction to whole-human  ID-based instruction by removing clothing-related phrases, retaining non-wearable objects, and remapping their placeholder IDs to whole-level identifiers. This yields two synchronized supervision formats: in the \textbf{part-human} format, all belongings remain as separate placeholders and the instruction uses \(c_{\mathrm{id}}^{\mathrm{part}}\); in the \textbf{whole-human} format, clothing references are absorbed into the human reference and the instruction uses \(c_{\mathrm{id}}^{\mathrm{whole}}\). This paired design decouples fine-grained appearance grounding from holistic identity composition. For the supplementary MICo-150K De\&Re subset, each case is converted into the same placeholder-based schema to enable joint training.

% The final dataset contains about 10k single-person samples, 5k multi-person samples, and 11k adapted MICo De\&Re samples, totaling 26k structured training samples. Table\ref{tab:dataset_overview} summarizes the reference types and supervision signals provided by the dataset.

\begin{figure*}[ht]
    \centering
    \includegraphics[width=\textwidth]{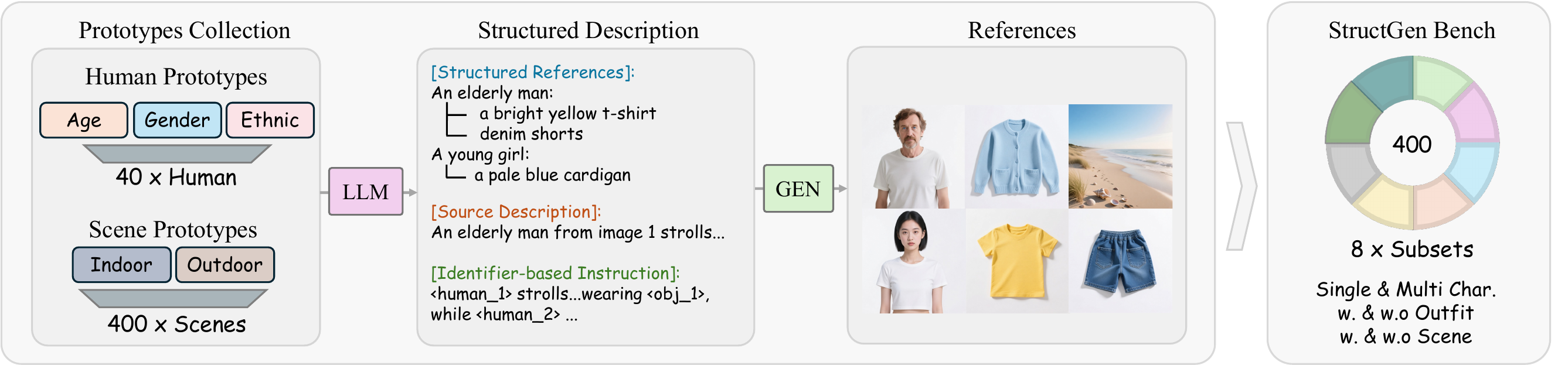}
    \vspace{-2em}
    \caption{Overview of StructGen Bench. The benchmark is organized along three factors: task type, subject number, and scene reference, resulting in eight subcategories. For each case, we first define human and scene prototypes, then synthesize complete descriptions and reference images for all entities, and construct both natural-language and identifier-based instructions.}
    \vspace{-1em}
    \label{fig:bench}
\end{figure*}

\subsection{Structured Tuning and Evaluation}
\label{sec:3_3}

\myparagraph{Structured Tuning}
Based on the dataset constructed above, we conducted the training of our StructGen within the BAGEL framework.
As illustrated in~\cref{fig:method}, we employ a simple yet effective training setting. The identifier for each entity in the reference dictionary, alongside the identifier-based instruction, are encoded by the text tokenizer, and the corresponding reference images are encoded by the image tokenizer. These components collectively constitute the structured context.
The model is trained to adapt to this new context format and correctly modeling it by minimizing the flow matching loss with the target image.

To further enhance the model's robustness and to support diverse application formats, we introduce the following mixed-sampling strategy during training.
Firstly, each training sample is randomly instantiated in either an Outfit-Conditioned or Outfit-Retaining format, enabling the model to jointly learn fine-grained attribute associating and holistic character composition.
Secondly, each references is also independently sampled between either the image or text modality with equal probability.
When text is sampled, the corresponding reference image is replaced by its detailed caption from our data pipeline and encoded by the model’s text branch. Consequently, our structured context framework can seamlessly handle generation based on both multi-reference images and multi-field detailed captions, as shown in~\cref{fig:t2i_generation}.

% encouraging the model to ground references through both visual and semantic signals.
% At the \emph{tas level}, each sample is randomly instantiated as either a part-human or whole-human version, so the model learns both fine-grained grounding and holistic composition. At the \emph{modality level}, each placeholder entry in \(\mathcal{D}_{\mathrm{ref}}\) is independently assigned to the image modality or the text modality with equal probability, encouraging the model to learn both visual and semantic reference grounding.

% \myparagraph{Mixed Sampling Strategy.}
% During training, we apply a mixed sampling strategy along two dimensions. 

\myparagraph{StructGen Bench}
Existing multi-reference benchmarks, such as OmniContext~\cite{wu2025omnigen2}, typically involve a limited number of reference images and contain relatively few human-centric scenarios. As a result, they are insufficient to comprehensively evaluate model performance in complex multi-reference settings, particularly when rich interactions between human and their surroundings are involved. Therefore, we establish a new benchmark that specifically targets challenging human-centric multi-reference generation.

Our benchmark is designed along three orthogonal factors: the task type (outfit-conditioned \vs outfit-retaining), the number of human subjects (single-human \vs multi-human), and the presence of background scene reference (with scene \vs without scene), resulting in 8 subsets and enabling evaluation across varying application scenarios.
The input configuration of each subset is illustrated in~\cref{fig:qualitative}.
An overall pipeline for constructing our benchmark is shown in~\cref{fig:bench}.
To enhance the balance and diversity of references in the benchmark, we first define numerous human and scene prototypes covering different attributes such as gender, age, ethnicity, and environments.
Subsequently, we prompt GPT to reasonably synthesize complete descriptions based on these prototypes, including the human’s actions, states, and belongings.
We further generate reference images corresponding through the synthesized descriptions using Qwen-Image.
To evaluate both existing models and our StructGen, we synthesize two versions of the instructions, corresponding to the pure natural language format used in prior works and our proposed identifier-based format.
All these components forms a complete sample in our benchmark.

To assess model performance, we focus on three key aspects: prompt following, subject consistency and facial ID consistency.
Prompt following (\textbf{PF}) assesses whether the generated image satisfies the intended composition, with particular emphasis on correctly modeling the subject–object associations.
Subject consistency (\textbf{SC}) evaluates whether the generated image faithfully preserves the referenced objects and scene, focusing on detailed consistency of attributes like color, material, and shape.
These two metrics are assessed through a VLM-as-a-judge~\cite{ku2024viescore,pengdreambench++,wu2025omnigen2, wu2026micon, huammig} paradigm with GPT-4.1.
Facial ID consistency (\textbf{ID}) is employed to measure whether the generated human faces keep consistent facial ID with references. We use InsightFace to extract facial ID embedding and Hungarian algorithm to perform one-to-one matching between references and generation.
The metric is then calculated as the sum of ID embedding similarities for all matched face pairs divided by the maximum faces count in the references and the generated image.
This formulation not only assesses facial similarity but also effectively penalizes semantic errors where the number of generated individuals does not match the reference count.

\section{Experiments}
\label{sec:exp}

\subsection{Experimental Setups}

\myparagraph{Implementation Details}
We train StructGen on 4 NVIDIA H20 GPUs with the model weights initialized from BAGEL-MICo.
For training efficiency, we only update the attention projection and normalization layers in both understanding and generation branches and keep all other parameters frozen.
To mitigate the training instability caused by the distribution gap between our real-world dataset and the initial weight of the model, we also incorporated a portion of the MICo data into our training process and train the model for approximately 13k steps. Please refer to our supplementary material for further details.
% We use a constant learning rate of \(2\times10^{-5}\) with 200 warmup steps and train for approximately 13k steps on the full StructGen Dataset 26k samples.
% Following ~\cref{sec:3_2}, each placeholder is independently grounded by either a reference image or a structured text description with equal probability, and part-human \& whole-human versions are sampled with equal probability.

\myparagraph{Comparing Methods and Benchmarks}
We compare StructGen with four representative baselines, including two pre-trained models, OmniGen2~\cite{wu2025omnigen2} and BAGEL~\cite{deng2025emerging}, as well as two distillation-based fine-tuned models, BAGEL-MICo~\cite{wei2025mico} and Echo-4o~\cite{ye2025echo}.
Evaluation is conducted on two benchmarks. (1)~\textbf{OmniContext}~\cite{wu2025omnigen2}: we use its human-centric subsets for comparison on public benchmarks. The evaluation metrics follow the original OmniContext protocol, including prompt following (PF) and subject consistency (SC). In addition, we introduce a Facial ID consistency (ID) metric, computed using InsightFace, to specifically assess identity consistency of human subjects. For compatibility with our framework, we construct a structured variant of OmniContext by equipping references and instructions with identifiers.
(2)~\textbf{StructGen Bench}: the evaluation metrics follow the definitions in~\cref{sec:3_3}. Input formats are adapted to each model: StructGen operates on structured inputs, while baseline methods use the natural language format.

\begin{table*}[t]
    \centering
    \small % 25列数据必须使用较小字号以防溢出
    \setlength{\tabcolsep}{1.6pt} % 极限压缩列间距
    \caption{Quantitative comparison on human-related subsets from OmniContext, with prompt following (PF), subject consistency (SC), overall scores, and facial ID similarity (ID). The best and second best results are highlighted in \textbf{Bold} and \underline{underline}.}
    \vspace{-1em}
    \begin{tabular}{l cccc cccc cccc cccc cccc cccc}
        \Xhline{0.8pt}
        \multirow{3}{*}{Method} & \multicolumn{4}{c}{Single Human}
        & \multicolumn{4}{c}{Multi Human}
        & \multicolumn{4}{c}{Human + Object}
        & \multicolumn{4}{c}{Human + Scene}
        & \multicolumn{4}{c}{Human + Obj. + Sce.}
        & \multicolumn{4}{c}{Average} \\
        \cmidrule(lr){2-5} \cmidrule(lr){6-9} \cmidrule(lr){10-13} \cmidrule(lr){14-17} \cmidrule(lr){18-21} \cmidrule(lr){22-25}
        & PF & SC & Over. & ID
        & PF & SC & Over. & ID
        & PF & SC & Over. & ID
        & PF & SC & Over. & ID
        & PF & SC & Over. & ID
        & PF & SC & Over. & ID \\
        \Xhline{0.5pt}
        OmniGen2~\cite{wu2025omnigen2}
        & 8.04 & 8.34 & 8.05 & 0.32
        & 7.70 & 6.96 & 7.11 & 0.15
        & 7.56 & 7.56 & 7.45 & 0.24
        & 7.06 & 5.94 & 6.38 & 0.18
        & \underline{7.50} & 6.68 & 7.04 & 0.21
        & 7.57 & 7.10 & 7.21 & 0.22 \\

        BAGEL~\cite{deng2025emerging}
        & 7.72 & 4.86 & 5.48 & 0.13
        & 6.14 & 4.86 & 5.17 & 0.18
        & 6.74 & 6.28 & 6.24 & 0.30
        & 4.56 & 3.94 & 4.07 & 0.21
        & 5.90 & 5.30 & 5.47 & 0.24
        & 6.21 & 5.05 & 5.29 & 0.21 \\

        BAGEL-MICo~\cite{wei2025mico}
        & \underline{8.38} & 8.26 & 8.15 & \underline{0.33}
        & 7.58 & 7.20 & 7.20 & 0.29
        & 7.62 & 8.28 & 7.88 & \underline{0.56}
        & 7.72 & 6.72 & 7.11 & 0.35
        & 6.98 & 7.16 & 7.02 & \underline{0.45}
        & 7.66 & 7.52 & 7.47 & \underline{0.40} \\

        Echo-4o~\cite{ye2025echo}
        & 8.14 & \underline{8.92} & \underline{8.47} & 0.26
        & \underline{7.78} & \underline{8.82} & \underline{8.15} & \underline{0.30}
        & \underline{7.78} & \underline{8.50} & \underline{8.11} & 0.43
        & \underline{7.92} & \underline{8.00} & \underline{7.82} & \underline{0.37}
        & \textbf{7.86} & \textbf{7.94} & \textbf{7.86} & 0.43
        & \underline{7.90} & \underline{8.44} & \underline{8.08} & 0.36 \\

        \textbf{StructGen (ours)}
        & \textbf{8.54} & \textbf{9.36} & \textbf{8.89} & \textbf{0.45}
        & \textbf{8.08} & \textbf{8.94} & \textbf{8.40} & \textbf{0.43}
        & \textbf{7.88} & \textbf{8.70} & \textbf{8.24} & \textbf{0.62}
        & \textbf{8.66} & \textbf{8.30} & \textbf{8.45} & \textbf{0.41}
        & 7.22 & \underline{7.54} & \underline{7.29} & \textbf{0.58}
        & \textbf{8.08} & \textbf{8.57} & \textbf{8.25} & \textbf{0.50} \\
        \Xhline{0.8pt}
    \end{tabular}
    \label{tab:omni_context_merged_all_metrics}
\end{table*}

\begin{figure*}[ht]
    \centering
    \includegraphics[width=\textwidth]{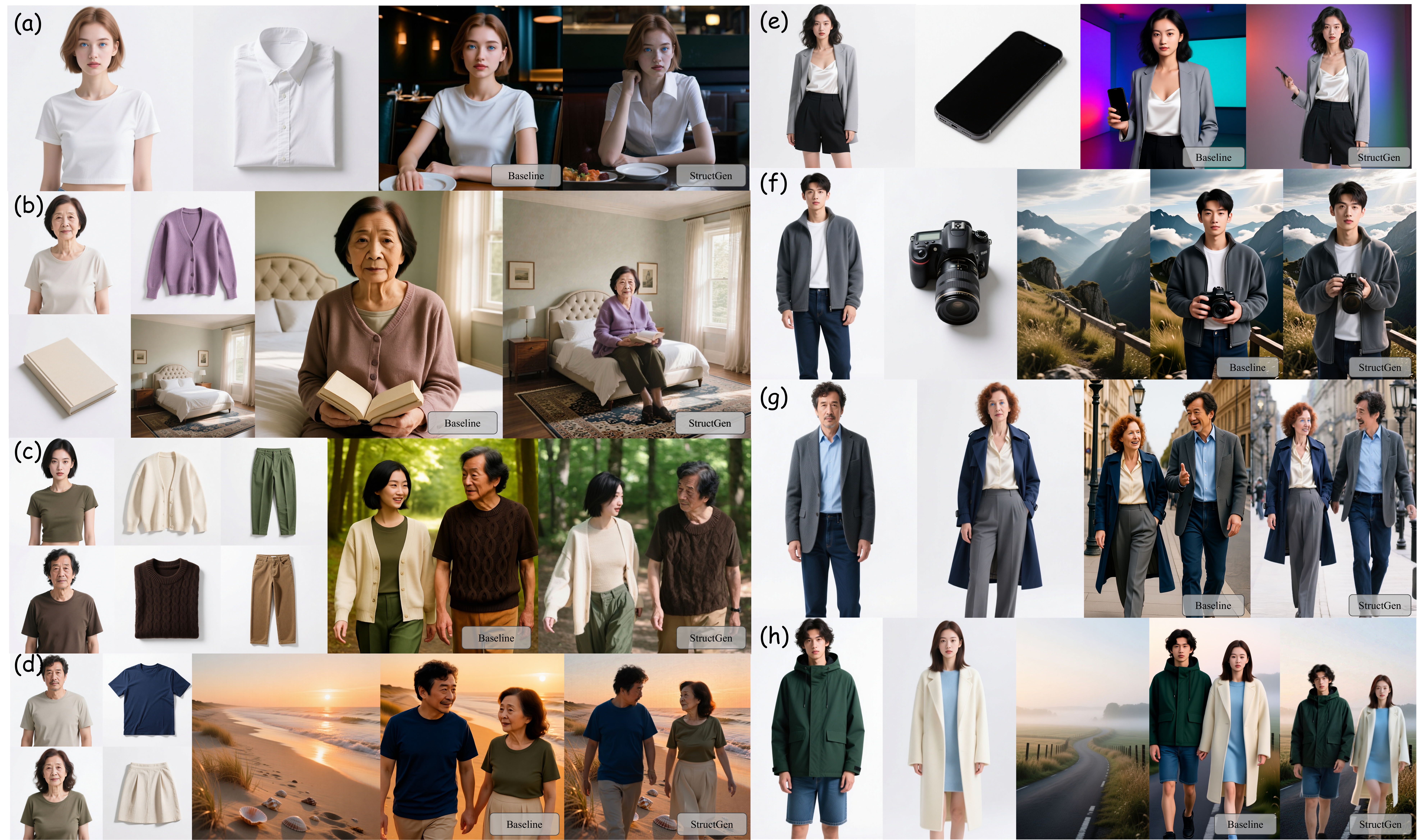}
    \vspace{-2em}
    \caption{Qualitative comparison on our proposed benchmark. StructGen demonstrates advantages on the consistency of human ID and background, the flexibility of human-environment interactions, and the realism of textures and lighting.}
    \vspace{-0.5em}
    \label{fig:qualitative}
\end{figure*}

\subsection{Experimental Results}

\begin{figure*}[ht]
    \centering
    \includegraphics[width=0.98\textwidth]{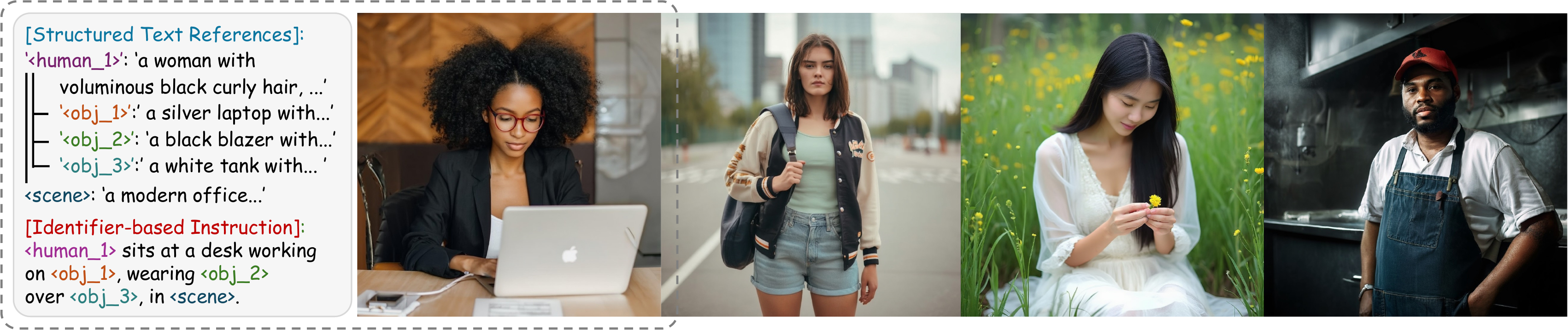}
    \vspace{-1em}
    \caption{StructGen performs generation with all references provided as textual descriptions, demonstrating the potential of structured context for text-to-image generation with decoupled entity descriptions and global semantic specification.}
    \label{fig:t2i_generation}
\end{figure*}

\myparagraph{Qualitative Results}
\cref{fig:qualitative} presents the qualitative comparison between our StructGen and the baseline model, Echo-4o, on our proposed benchmark.
The left columns (a)–(d) present examples of the Outfit-Conditioned task across single- and multi-human settings, with and without scene references, while the right columns correspond to the Outfit-Retaining task.
Compared to the baseline, StructGen achieves better consistency in objects (\eg, the white shirt in (a)) and facial identity (\eg, the young man in (f)), and produces more natural subject–environment interactions with improved background maintaining, as shown in (b). 
In addition, across all generated images, StructGen consistently exhibits more natural and realistic visual details, such as lighting and texture. This underscores another advantage of our real-world image-based data curation strategy.
\cref{fig:t2i_generation} presents the generation results of StructGen where all reference fields are provided as textual descriptions instead of images. This highlights an additional potential of our structured framework: decoupling the entity descriptions from global semantic specification, thereby enabling text-to-image generation with both detailed individual attributes and precise inter-entity associations.

\myparagraph{Quantitative Results}
\Cref{tab:omni_context_merged_all_metrics} presents results on OmniContext benchmark. StructGen ranks first in terms of both the average metrics and the vast majority of individual metrics. 
The advantage of StructGen is especially pronounced on challenging compositional subsets, \eg, on \textit{Human + Scene} subset, StructGen exhibits a significant improvement in semantic alignment (PF), showing its capability to generate coherent integration of human and background environment.
% demonstrating that the reference dictionary effectively prevents misassigning attributes to wrong subjects, such as swapping clothing or accessories between characters.
Furthermore, StructGen achieves consistent and remarkable gains in facial identity consistency (ID).
It reaches an average score of 0.50, surpassing the second-best method by 25\%, and the gap widens to over 43\% on \textit{Multi Human} subset.
% These results show that the structured context and the high-quality training data substantially improve the semantic alignment and identity degradation and appearance inconsistency commonly observed in existing methods.

\begin{table*}[t]
    \centering
    \small % 建议使用 small 字体以适应 16 列的宽度
    \setlength{\tabcolsep}{7pt} % 进一步压缩列间距
    \caption{Quantitative comparison on our proposed benchmark. The best and second best results are highlighted in \textbf{Bold} and \underline{underline}. $^\star$OmniGen2 supports at most 5 reference images, subsets exceeding this threshold are omitted.}
    \vspace{-1em}
    \begin{tabular}{l ccc ccc ccc ccc ccc}
        \Xhline{0.8pt}
        \multirow{3}{*}{Method} & \multicolumn{3}{c}{Single Human}
        & \multicolumn{3}{c}{Single Human + Scene}
        & \multicolumn{3}{c}{Multi Human}
        & \multicolumn{3}{c}{Multi Human + Scene}
        & \multicolumn{3}{c}{Average} \\
        \cmidrule(lr){2-4} \cmidrule(lr){5-7} \cmidrule(lr){8-10} \cmidrule(lr){11-13} \cmidrule(lr){14-16}
        & PF & SC & ID & PF & SC & ID & PF & SC & ID & PF & SC & ID & PF & SC & ID \\
        \Xhline{0.5pt}
        \multicolumn{16}{l}{\textit{Task A: Outfit-Conditioned}} \\
        % \Xhline{0.3pt}
        OmniGen2$^\star$~\cite{wu2025omnigen2} & 6.30 & 8.64 & 0.45 & 6.38 & 8.73 & 0.42 & -- & -- & -- & -- & -- & -- & -- & -- & -- \\
        BAGEL~\cite{deng2025emerging} & 6.50 & 8.68 & 0.39 & 6.72 & 8.56 & 0.34 & 6.82 & 8.33 & 0.10 & 6.34 & 8.09 & 0.13 & 6.59 & 8.41 & 0.24 \\
        BAGEL-MICo~\cite{wei2025mico} & \underline{6.62} & \underline{8.94} & \underline{0.61} & 6.80 & \textbf{8.83} & \textbf{0.55} & 7.20 & \underline{8.52} & 0.22 & 6.48 & 8.41 & 0.26 & \underline{6.78} & \underline{8.68} & \underline{0.41} \\
        Echo-4o~\cite{ye2025echo} & 6.20 & 8.75 & 0.42 & \underline{6.84} & 8.73 & 0.45 & \underline{7.32} & 8.37 & \underline{0.29} & \underline{6.68} & \underline{8.48} & \underline{0.27} & 6.76 & 8.58 & 0.36 \\
        \textbf{StructGen (ours)} & \textbf{6.90} & \textbf{9.05} & \textbf{0.71} & \textbf{7.88} & \underline{8.78} & \underline{0.47} & \textbf{8.14} & \textbf{8.61} & \textbf{0.36} & \textbf{7.72} & \textbf{8.59} & \textbf{0.32} & \textbf{7.66} & \textbf{8.76} & \textbf{0.47} \\
        \Xhline{0.5pt}
        \multicolumn{16}{l}{\textit{Task B: Outfit-Retaining}} \\
        % \Xhline{0.3pt}
        OmniGen2$^\star$~\cite{wu2025omnigen2} & \underline{6.58} & \underline{9.19} & 0.36 & 7.30 & \textbf{9.00} & 0.32 & -- & -- & -- & -- & -- & -- & -- & -- & -- \\
        BAGEL~\cite{deng2025emerging} & 5.68 & 8.25 & 0.42 & 5.78 & 8.26 & 0.26 & 5.76 & 6.03 & 0.22 & 4.52 & 5.91 & 0.17 & 5.43 & 7.11 & 0.27 \\
        BAGEL-MICo~\cite{wei2025mico} & 6.40 & 9.00 & \underline{0.46} & 6.88 & \underline{8.78} & \textbf{0.46} & 6.28 & 6.45 & 0.27 & 5.24 & \underline{6.43} & \textbf{0.27} & 6.20 & 7.67 & \underline{0.37} \\
        Echo-4o~\cite{ye2025echo} & 6.04 & 8.75 & 0.39 & \underline{7.54} & 8.70 & 0.38 & \underline{6.32} & \underline{7.46} & \underline{0.29} & \underline{5.50} & 6.20 & 0.26 & \underline{6.35} & \underline{7.78} & 0.33 \\
        \textbf{StructGen (ours)} & \textbf{6.66} & \textbf{9.25} & \textbf{0.59} & \textbf{7.78} & 8.74 & \textbf{0.46} & \textbf{7.18} & \textbf{8.16} & \textbf{0.40} & \textbf{6.80} & \textbf{6.66} & \textbf{0.27} & \textbf{7.11} & \textbf{8.20} & \textbf{0.43} \\
        \Xhline{0.8pt}
    \end{tabular}
    \label{tab:combined_benchmark}
\end{table*}

% \myparagraph{Quantitative Results on StructGen Bench}
\Cref{tab:combined_benchmark} shows the results on our proposed StructGen Bench.
As OmniGen2 supports at most 5 reference images, results on subsets that exceed this limit are therefore omitted.
StructGen leads in average prompt following by +0.88 on Outfit-Conditioned task and +0.76 on Outfit-Retaining task over the respective second-best baselines.
On the challenging \textit{Multi Human + Scene} subset, BAGEL drops to only 4.52 and Echo-4o manages only 5.50, both essentially failing to produce semantically correct compositions, whereas StructGen still scores 6.80. This confirms that the structured framework performs robustly with increasing association complexity.
StructGen also exhibit clear advantage in generation consistency of facial ID and objects, achieving the first position in terms of average ID and SC metrics for both two tasks.
On the most challenging subsets, baseline method drop significantly to 0.13, nearly unable to preserve any consistency in identity, whereas our StructGen still maintains a robust performance of 0.36.
% BAGEL collapses to merely 0.10 ID on Outfit-Conditioned, nearly unable to preserve any identity, while StructGen maintains 0.36.
% For objects consistency, StructGen achieves 8.20 on Outfit-Retaining compared to Echo-4o's 7.78, and the gap reaches +0.70 on Multi Human of Outfit-Retaining. This advantage reflects that our real-image-based training data captures richer appearance details than distillation-based datasets. The latter tend to produce close-up views with limited background context, resulting in degraded fine-grained fidelity for both objects and scenes.
% Note that OmniGen2 supports at most 5 reference images; multi-character subsets that exceed this limit are therefore omitted for this method.

\begin{table}[t]
    \centering
    \small
    \caption{Ablation studies of different configurations.}
    \vspace{-1em}
    \label{tab:ablation}
    \setlength{\tabcolsep}{3.5pt}
    \begin{tabular}{lccccc}
        \Xhline{0.8pt}
        Configuration & (a) & (b) & (c) & (d) & (e) \\
        \Xhline{0.5pt}
        % ID: Single/Multi & 0.43/0.21 & 0.46/0.19 & 0.37/0.23 & 0.51/0.30 & \textbf{0.65}/\textbf{0.38} \\
        ID: Single/Multi & 0.51/0.30 & 0.37/0.23 & 0.43/0.21 & 0.46/0.19 & \textbf{0.65}/\textbf{0.38} \\
        \Xhline{0.8pt}
    \end{tabular}
    \vspace{-2em}
\end{table}

\subsection{Ablation Studies}
To evaluate the effectiveness of the structured context and new dataset, disentangle their individual contributions, and assess key design choices, we conduct a series of ablation studies, with results summarized in~\cref{tab:ablation}.
The evaluated configurations are as follows:
(a) training on our curated dataset using natural language format;
(b) training on MICo data while converted into our structured format;
(c) removing the post-filtering stage in the data curation pipeline;
(d) training only the generation branch while freezing the remaining parameters;
and (e) our final configuration. 
Each variant (a)–(d) isolates a single factor relative to the full configuration (e).
For evaluation, we select two challenging subsets from StructGen Bench as representatives, the single- and multi-human settings with scene references under the Outfit-Conditioned task, and report facial ID consistency (ID) results.

The results show that configuration (e) significantly outperforms both (a) and (b), validating that structured context modeling and our curated dataset each contribute independently to performance gains.
The improvement over (c) further highlights the importance of the post-filtering stage.
Finally, the comparison between (e) and (d) indicates that, although our method focuses on generation, fine-tuning parts of the understanding branch is crucial. This is consistent with our design, as the structured context requires the model to understand the new input format and correctly encode the identifiers and identifier-based instructions.

\section{Conclusion and Discussion}
\label{sec:conclu}
We present StructGen, a structured context modeling framework for human-centric multi-reference image generation. We show that the limitations of existing approaches largely arise from the ambiguity of free-form natural language instructions, which becomes increasingly verbose and unreliable as the number of references grows.
StructGen addresses this issue by organizing references into a dictionary-like format with unique identifiers, transforming implicit cross-reference associations into explicit grounding. This design avoids entangling content and relationships in ambiguous natural language, enabling more precise attribute--subject correspondence.
To support this formulation, we develop a structured data curation pipeline, preserving the semantic richness and spatial diversity of on real-world images, which is absent in prior distillation-based datasets.
We further establish a dedicated benchmark for challenging human-centric scenarios, with fine-grained evaluation of prompt following and generation consistency.
Extensive experiments on both public and our proposed benchmarks demonstrate that StructGen consistently outperforms existing methods, particularly in complex multi-reference settings involving multiple subjects and attribute–subject relationships.

\myparagraph{Limitation}
This work provides a preliminary exploration of structured context modeling and an extraction-based data curation pipeline. As an initial study, we focus on human-centric scenarios and construct a relatively small-scale fine-tuning dataset (approximately 16k samples) for validation. In future work, we plan to extend this framework to larger-scale, broader reference categories and more general multi-modal settings, aiming to further advance structured context modeling toward universal generation tasks.

\begin{acks}
This work was supported by the National Natural Science Foundation of China (No.92470203), National Natural Science Foundation of China (No.62472104), and Beijing Natural Science Foundation (No.L242022, L252025).
\end{acks}

\bibliographystyle{ACM-Reference-Format}
\bibliography{sample-base}

\clearpage
\appendix
\onecolumn

\begin{center}
    {\Huge\bfseries
    StructGen: Disambiguating Multi-Reference Image Generation
    via Structured Context Modeling}\\[0.4em]
    {\huge\normalfont Supplementary Material}
\end{center}

\vspace{0.5em}

\begin{multicols}{2}

\noindent This supplementary material provides additional details and visualizations for \textbf{StructGen}. 
Specifically, we include:
(1) detailed descriptions of the structured data curation pipeline and representative training data samples,
(2) additional details of the StructGen Bench construction pipeline and representative benchmark samples,
(3) implementation and training details,
and (4) more qualitative comparisons with baseline model.

\section{Data Curation Pipeline and Training Samples}

This section elaborates on the four-stage data curation pipeline introduced in the main paper, providing the detailed prompt designs for each stage and representative training samples.

\begin{figure*}[ht]
    \centering
    \includegraphics[width=\textwidth]{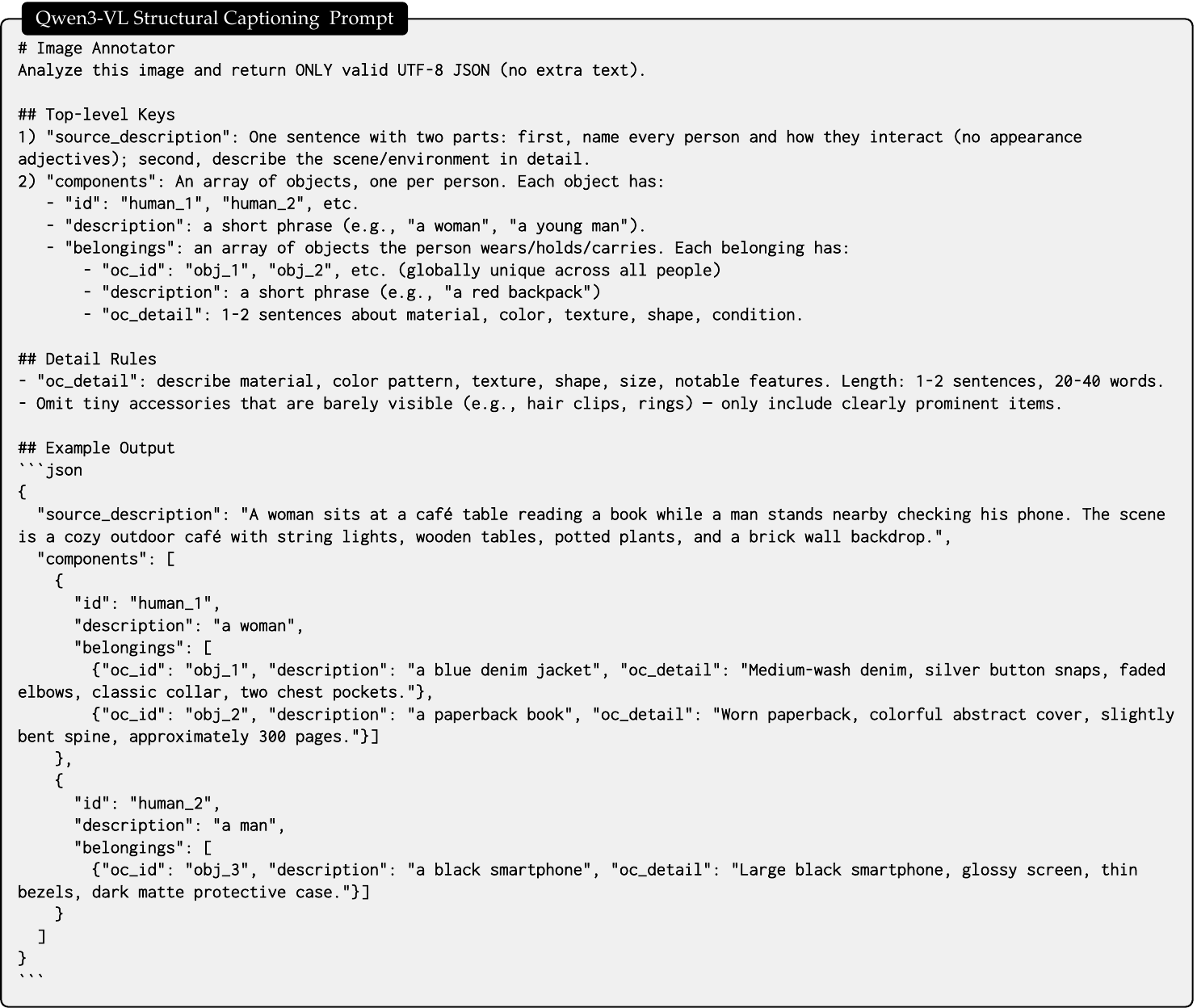}\\
    \vspace{0.2em}
    \includegraphics[width=\textwidth]{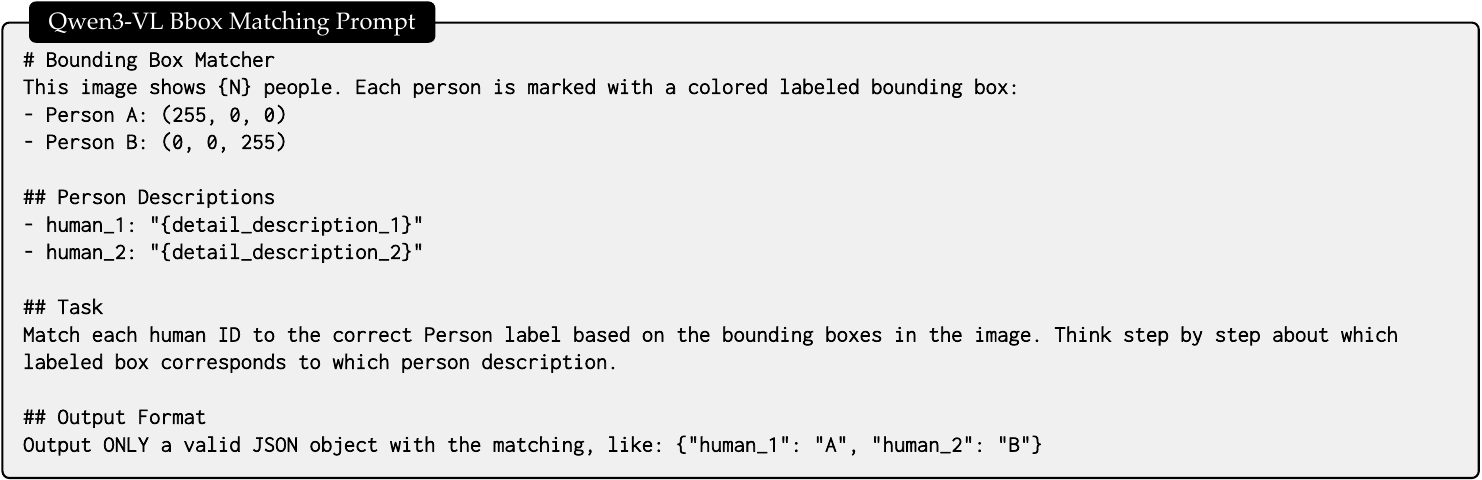}
    \vspace{-2em}
    \caption{\textbf{(Top)} Qwen3-VL~\cite{Qwen3-VL} produces structured captions (\textbf{oc\_id}: Outfit-Conditioned identifier; \textbf{oc\_detail}: detail description). \textbf{(Bottom)} VLM spatial matching prompt for multi-human images.}
    \label{fig:step1_vlm_captioning_prompt}
\end{figure*}

\myparagraph{Reference-Aware Captioning.}
For each source image, we employ Qwen3-VL~\cite{Qwen3-VL} to produce structured JSON annotations.
The annotation covers a comprehensive full-image caption capturing global compositional information and cross-entity associations, per-human component descriptions and per-belonging appearance attributes.
The prompt design and expected output format are illustrated in~\cref{fig:step1_vlm_captioning_prompt}.
For multi-human images, an additional VLM-based matching step is performed prior to extraction to establish spatial correspondence between each annotated \(\langle \mathrm{human}_i \rangle\) component and its detected bounding box.
Concretely, the VLM is presented with the original image annotated with colored bounding boxes (each labeled with a distinct letter identifier) alongside the textual descriptions of each human component, and outputs a JSON object matching each human identifier to the corresponding bounding box index.
The matching prompt is shown in~\cref{fig:step1_vlm_captioning_prompt}.

\begin{figure*}[ht]
    \centering
    \includegraphics[width=\textwidth]{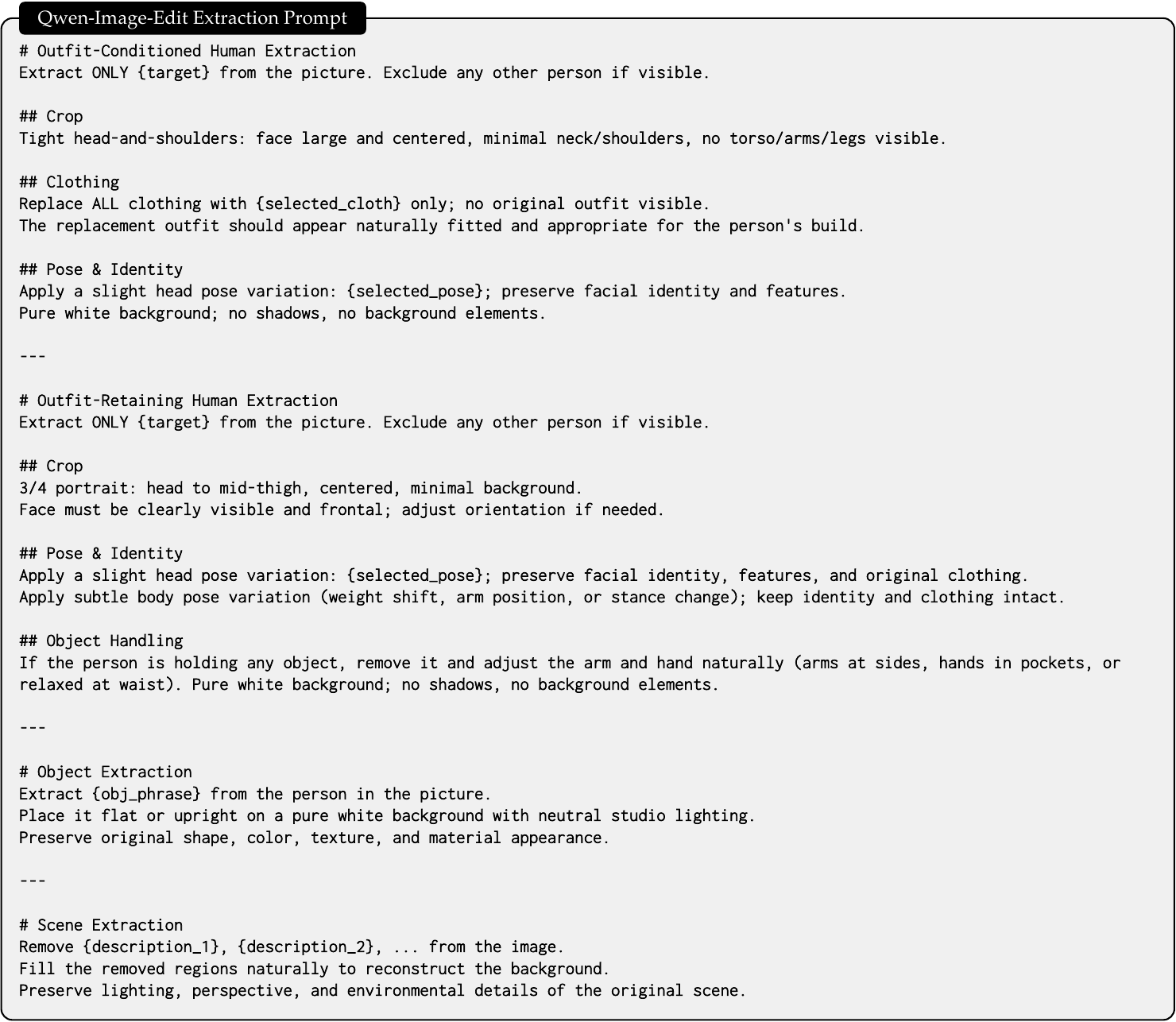}
    \caption{Reference Extraction prompts for all entity categories: Outfit-Conditioned human references (head-and-shoulders, outfit replaced), Outfit-Retaining human references (three-quarter-length, original outfit preserved), belonging references (white background), and scene references (foreground removed).}
    \label{fig:step3_gen_extract_prompt}
\end{figure*}

\myparagraph{Reference Extraction.}
Reference images for each entity are generated using Qwen-Image-Edit~\cite{wu2025qwen}.
For human references, the extraction follows two complementary task variants,
as illustrated in~\cref{fig:step3_gen_extract_prompt}.
In the Outfit-Conditioned variant, the reference is a head-and-shoulders
portrait with the original clothing replaced by a randomly sampled outfit from a
predefined catalog, covering: T-shirts, Shirts, Sweaters, Jackets, Hoodies, Dresses,
Tops, Bottoms, and Suits.
In the Outfit-Retaining variant, the reference is a three-quarter-length
portrait with the original outfit preserved intact.
In both variants, to prevent copy-paste bias, the head pose is
randomly sampled from the following candidates:
\begin{itemize}
% [leftmargin=*, noitemsep, topsep=2pt]
\item with the head slightly turned to the left
\item with the head slightly turned to the right
\item with the head slightly tilted
\item with the chin slightly raised
\item with the chin slightly lowered
\item with a subtle three-quarter face angle
\item with a slight profile face angle
\item looking slightly off-center
\end{itemize}
For multi-human images, to further avoid cross-person interference during extraction,
the extraction is applied to per-person cropped views obtained via the bounding box
matching step illustrated in~\cref{fig:step1_vlm_captioning_prompt}, rather
than on the full image directly.

\begin{figure*}[ht]
    \centering
    \includegraphics[width=\textwidth]{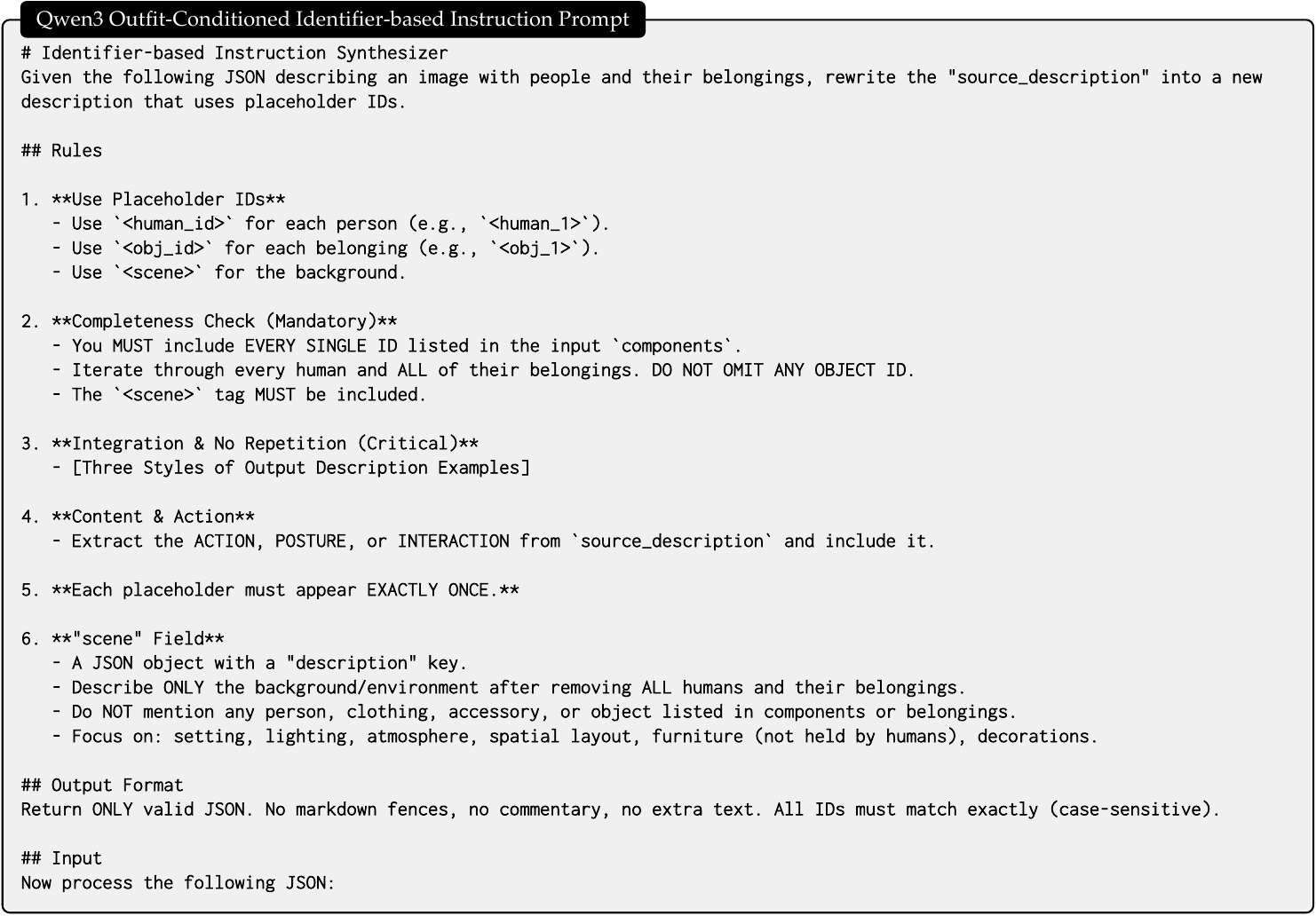}\\
    \vspace{0.3em}
    \includegraphics[width=\textwidth]{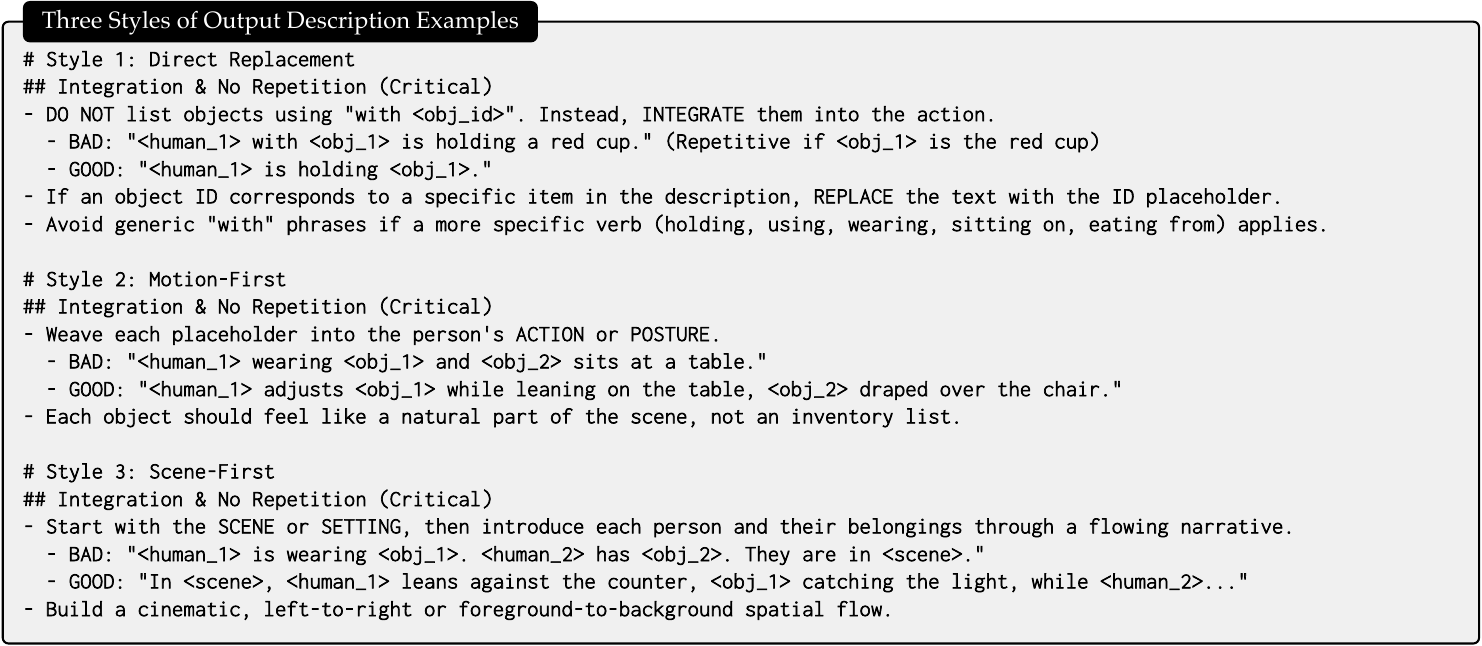}
    % \vspace{-2em}
    \caption{\textbf{(Top)} Outfit-Conditioned identifier-based instruction synthesis prompt. The LLM rewrites the full-image caption by replacing each entity mention with its corresponding placeholder identifier. \textbf{(Bottom)} Rewriting examples illustrating the three instruction styles: direct substitution, action-integrated, and narrative-flow.}
    \label{fig:step4_llm_outfit_conditioned_id_description_prompt}
\end{figure*}

\myparagraph{Identifier-based Instruction Synthesis.}
For the Outfit-Conditioned task, we employ Qwen3~\cite{yang2025qwen3} to
rewrite the full-image caption into an identifier-based instruction by replacing each
entity mention with its corresponding identifier.
To increase the lexical and syntactic diversity of training instructions, we adopt three
complementary writing styles: (1) \textit{direct substitution}, where each entity text
is replaced inline by its placeholder; (2) \textit{action-integrated}, where placeholders
are woven naturally into the subject's action or posture description; and (3)
\textit{narrative-flow}, where the description follows a scene-first spatial structure.
Each style is randomly assigned to individual training samples.
The detailed prompt for Outfit-Conditioned instruction synthesis is illustrated
in~\cref{fig:step4_llm_outfit_conditioned_id_description_prompt}, and representative
examples across all three styles are shown
in~\cref{fig:step4_llm_outfit_conditioned_id_description_prompt}.

\begin{figure*}[ht]
    \centering
    \includegraphics[width=\textwidth]{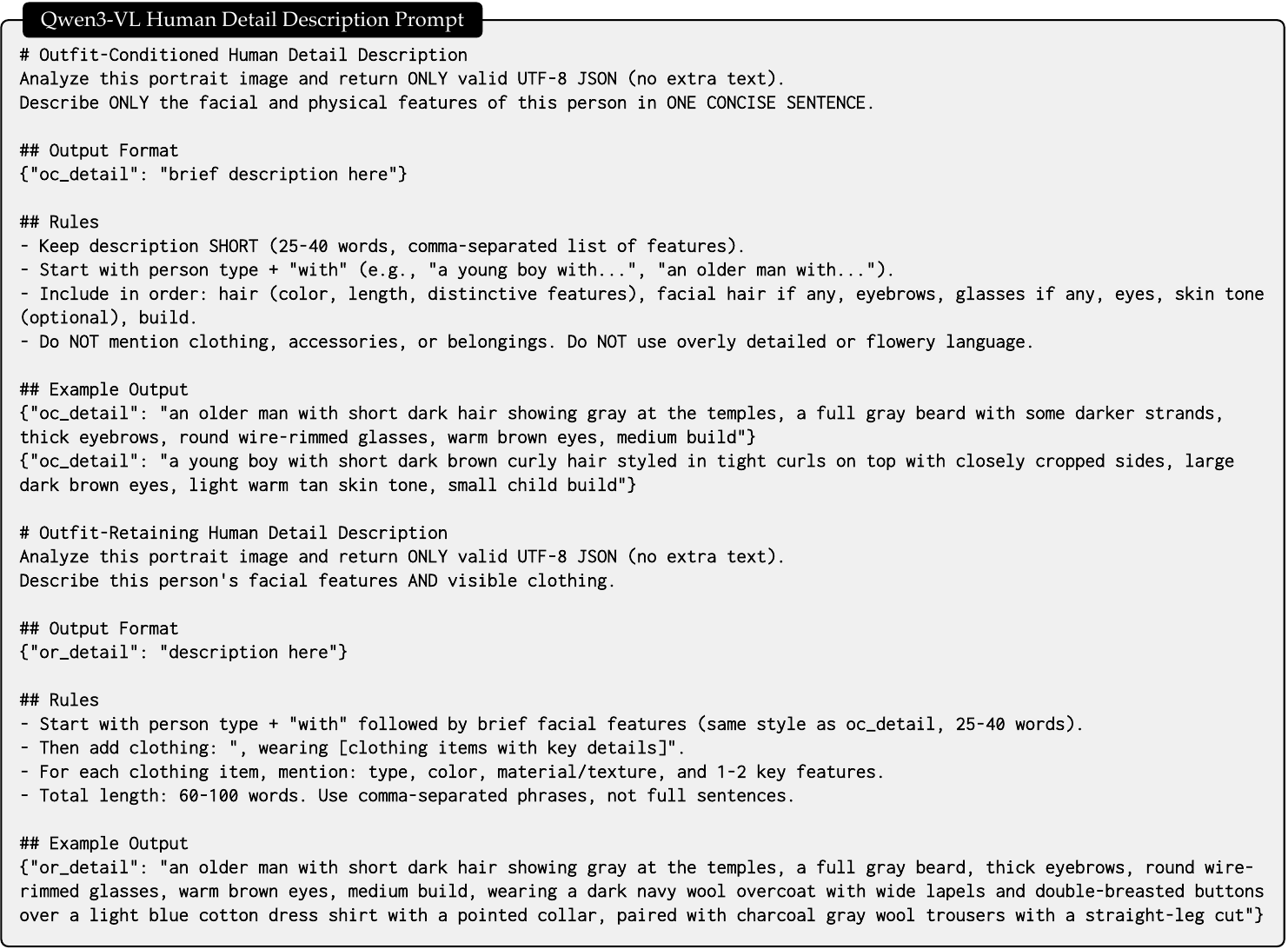}\\
    \vspace{0.3em}
    \includegraphics[width=\textwidth]{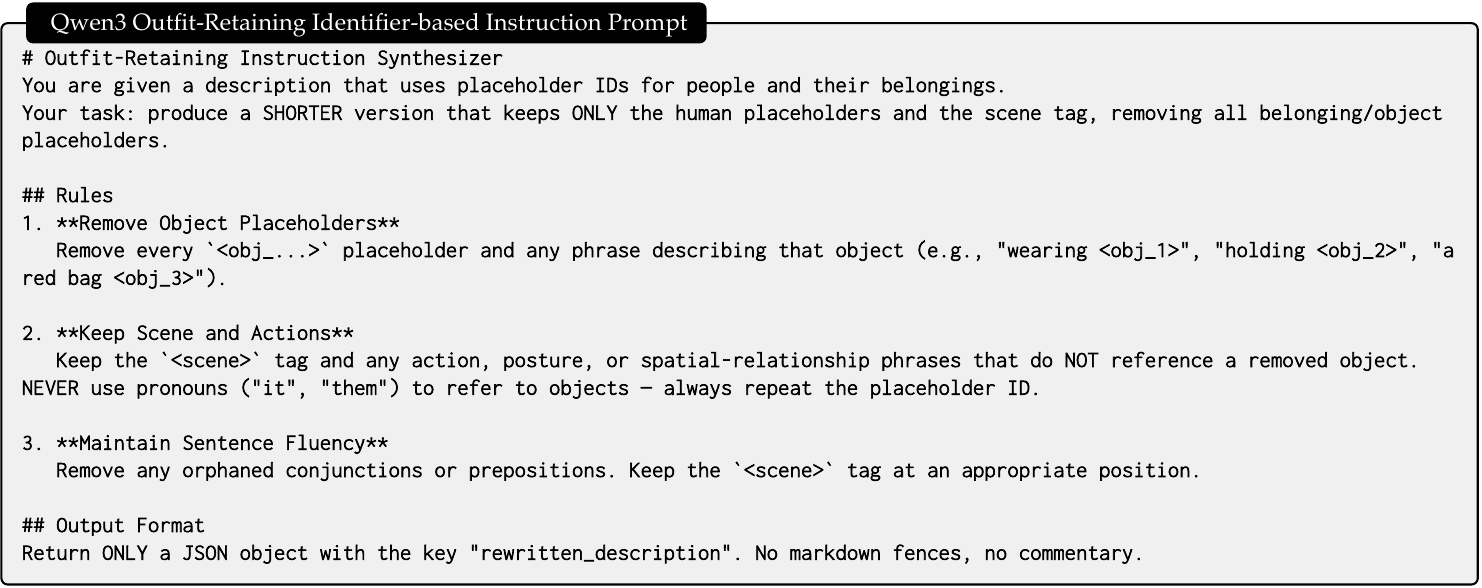}
    % \vspace{-2.5em}
    \caption{\textbf{(Top)} Qwen3-VL Human Detail Description Prompt. Qwen3-VL~\cite{Qwen3-VL}
    generates \textbf{oc\_detail} (facial features only) for Outfit-Conditioned references
    and \textbf{or\_detail} (facial features + clothing) for Outfit-Retaining references.
    Clothing classification is performed separately via keyword matching.
    \textbf{(Bottom)} Qwen3 Outfit-Retaining ID Description Prompt. The LLM removes
    clothing-related placeholders and remaps remaining object identifiers to their
    Outfit-Retaining counterparts.}
    \label{fig:step5_vlm_cloth_judge_and_human_detail_description_prompt}
\end{figure*}

In parallel, Qwen3-VL~\cite{Qwen3-VL} is applied to each extracted portrait reference
to generate fine-grained facial appearance descriptions and full-body appearance
descriptions that additionally include clothing details, and to determine whether each
belonging qualifies as a clothing item via keyword matching against a predefined cloth
vocabulary(\eg, shirts, jackets, dresses, trousers).
The generated appearance descriptions serve as text-modality reference inputs during
mixed-modality training, and the clothing classification results are subsequently used to
derive the Outfit-Retaining instructions.
The prompt for this step is shown
in~\cref{fig:step5_vlm_cloth_judge_and_human_detail_description_prompt}.

\begin{figure*}[ht]
    \centering
    \includegraphics[width=0.9\textwidth]{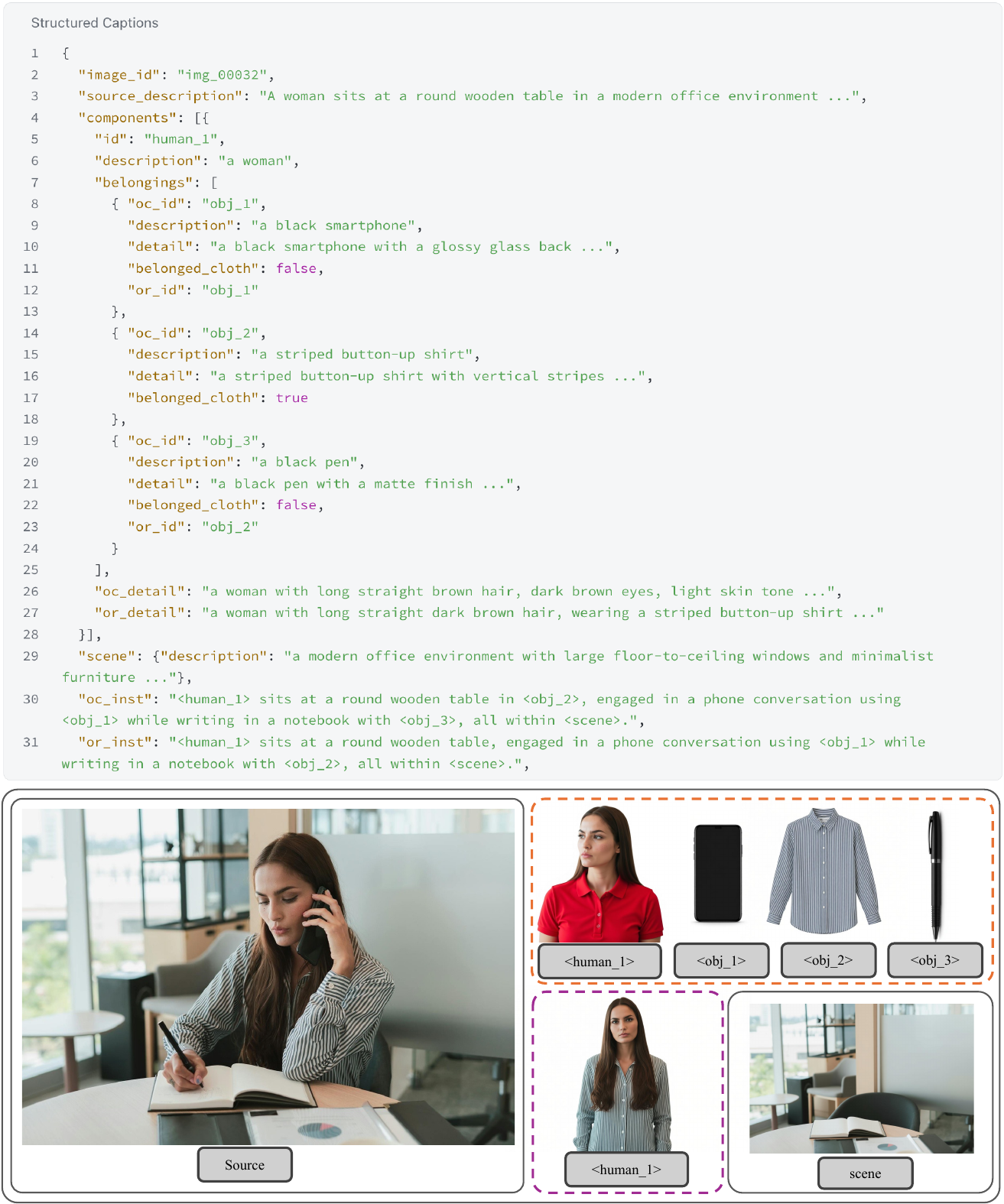}
    \caption{Training data sample 1: single-person example with human, belonging, and scene references. \textbf{oc\_id}: Outfit-Conditioned identifier; \textbf{or\_id}: Outfit-Retaining identifier; \textbf{oc\_inst}: Outfit-Conditioned instruction; \textbf{or\_inst}: Outfit-Retaining instruction; \textbf{oc\_to\_or\_map}: mapping from Outfit-Conditioned to Outfit-Retaining identifiers.}
    \label{fig:training_sample_single_1}
\end{figure*}

\begin{figure*}[ht]
    \centering
    \includegraphics[width=0.9\textwidth]{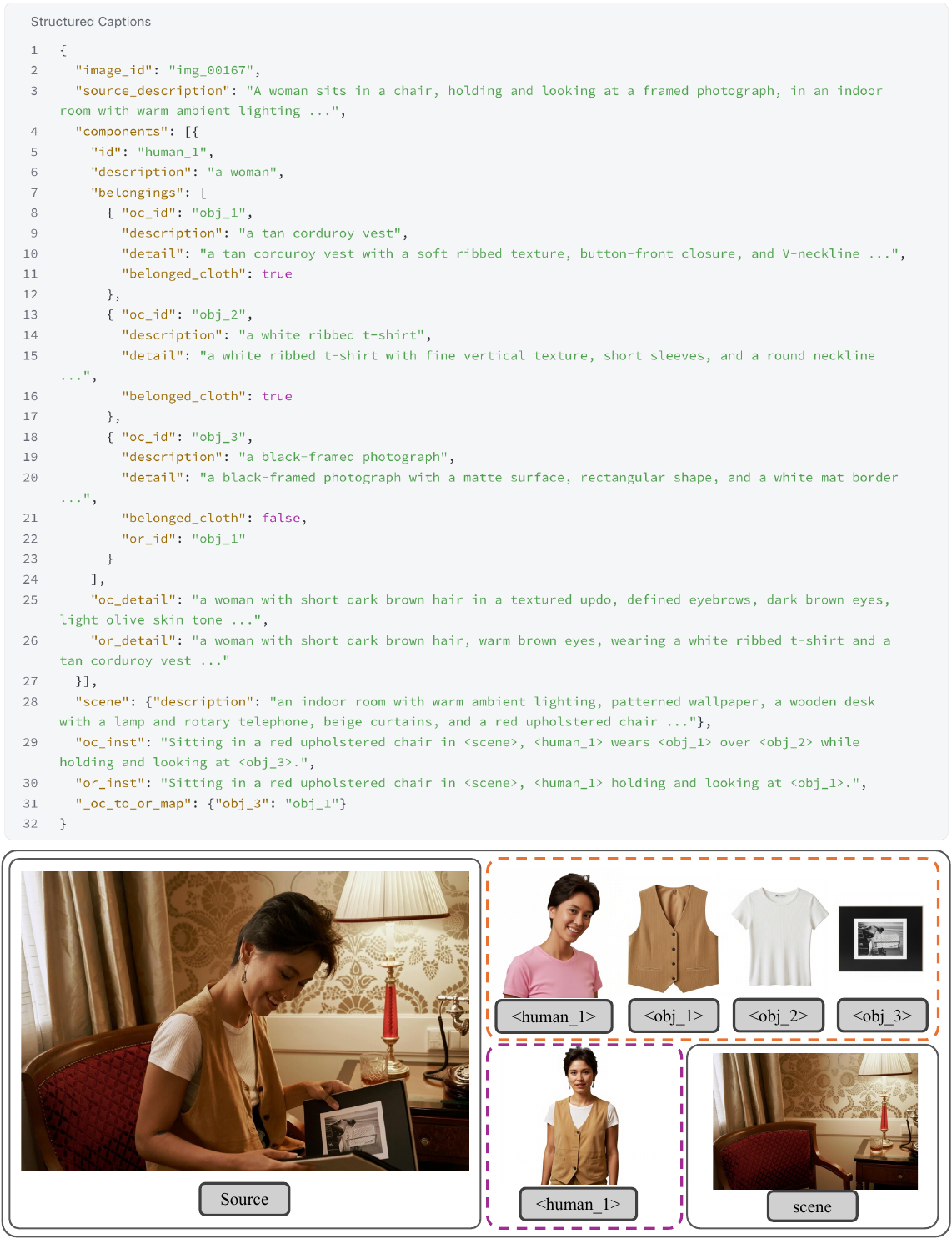}
    \caption{Training data sample 2: a second single-person example. Field names follow the same convention as in~\cref{fig:training_sample_single_1}.}
    \label{fig:training_sample_single_2}
\end{figure*}

\begin{figure*}[ht]
    \centering
    \includegraphics[width=0.9\textwidth]{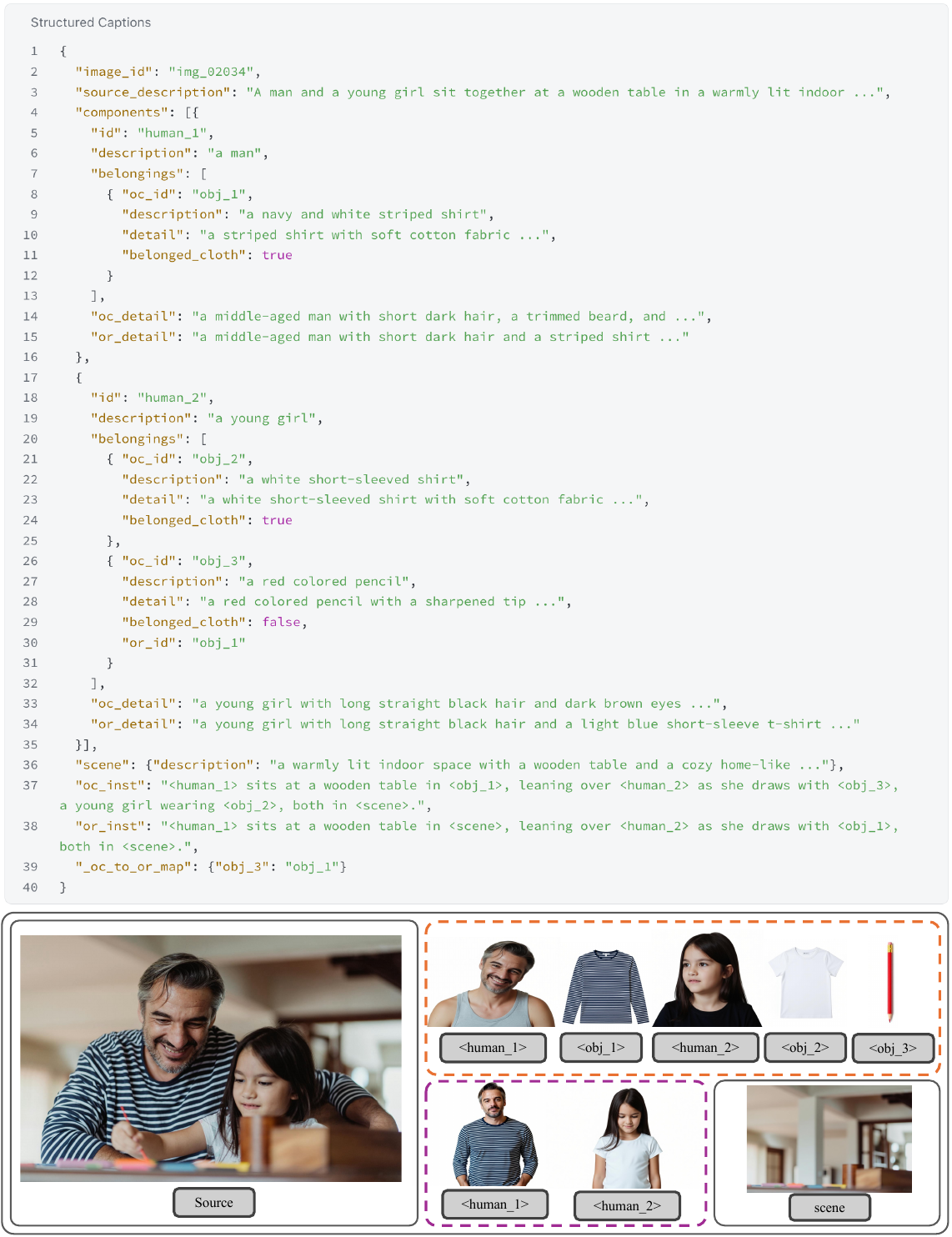}
    \caption{Training data sample 3: multi-person example with two human subjects and multiple belonging references. Field names follow the same convention as in~\cref{fig:training_sample_single_1}.}
    \label{fig:training_sample_multi_1}
\end{figure*}

\begin{figure*}[ht]
    \centering
    \includegraphics[width=0.9\textwidth]{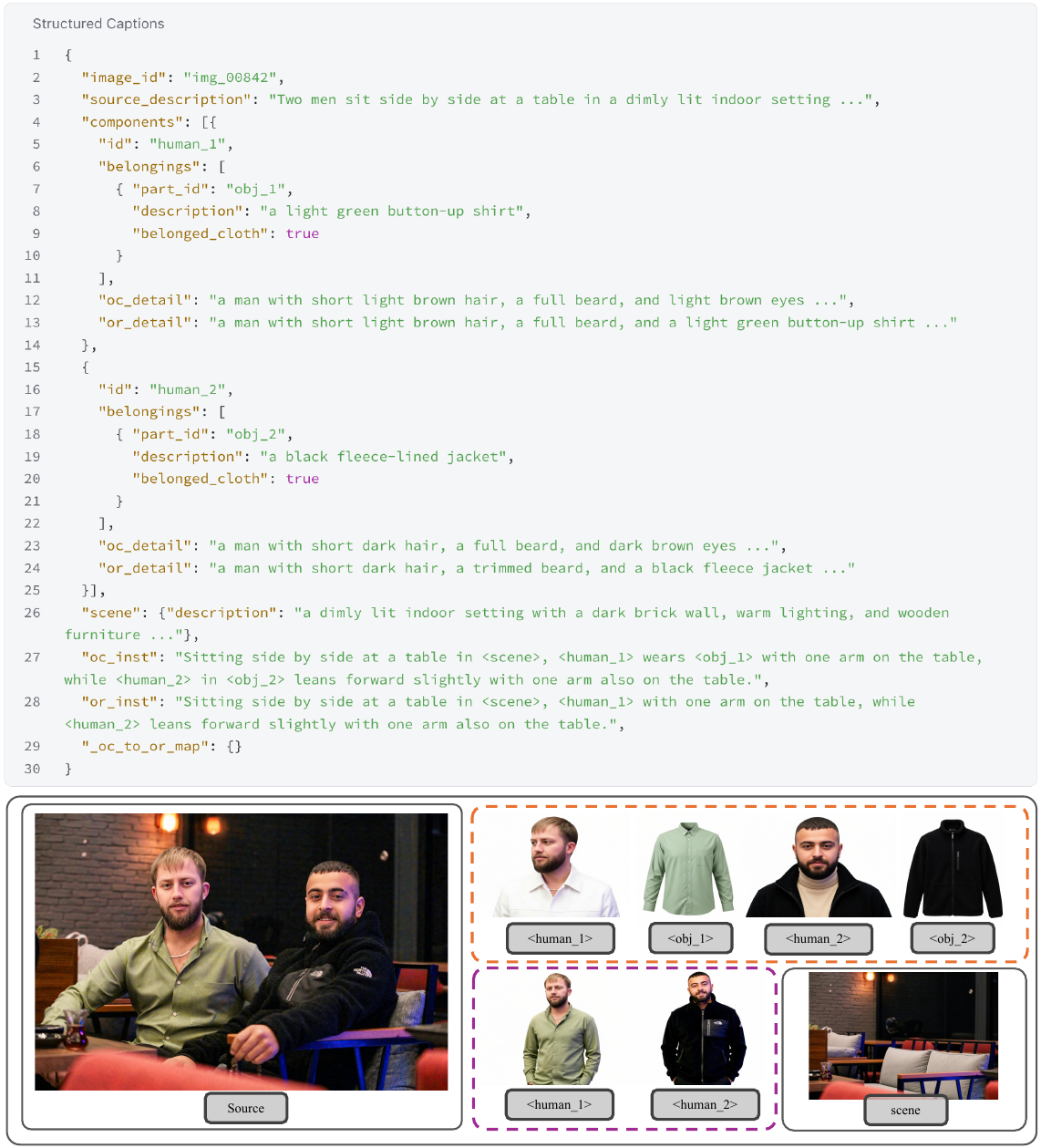}
    \caption{Training data sample 4: a second multi-person example. Field names follow the same convention as in~\cref{fig:training_sample_single_1}.}
    \label{fig:training_sample_multi_2}
\end{figure*}

For the Outfit-Retaining task, the identifier-based instruction is derived from
the Outfit-Conditioned instruction by removing all clothing-related placeholders and
remapping the remaining object identifiers to their whole-human counterparts.
The detailed prompt is shown
in~\cref{fig:step5_vlm_cloth_judge_and_human_detail_description_prompt}.
Representative training samples from the single- and multi-person subsets are
shown in Figures~\ref{fig:training_sample_single_1}--\ref{fig:training_sample_multi_2}.

\section{Benchmark Construction and Samples}

This section provides further details on the three-step construction pipeline of StructGen Bench, complementing the overview presented in the main paper.

\begin{figure*}[ht]
    \centering
    \includegraphics[width=\textwidth]{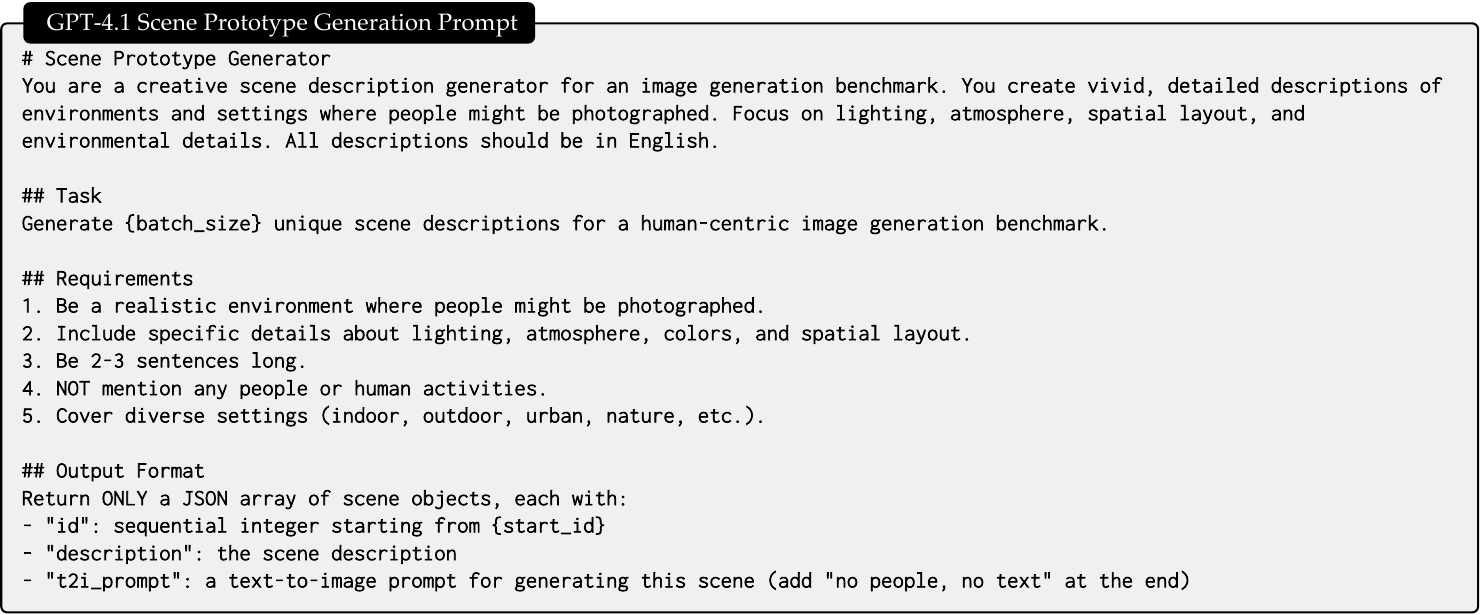}\\
    \vspace{0.2em}
    \includegraphics[width=\textwidth]{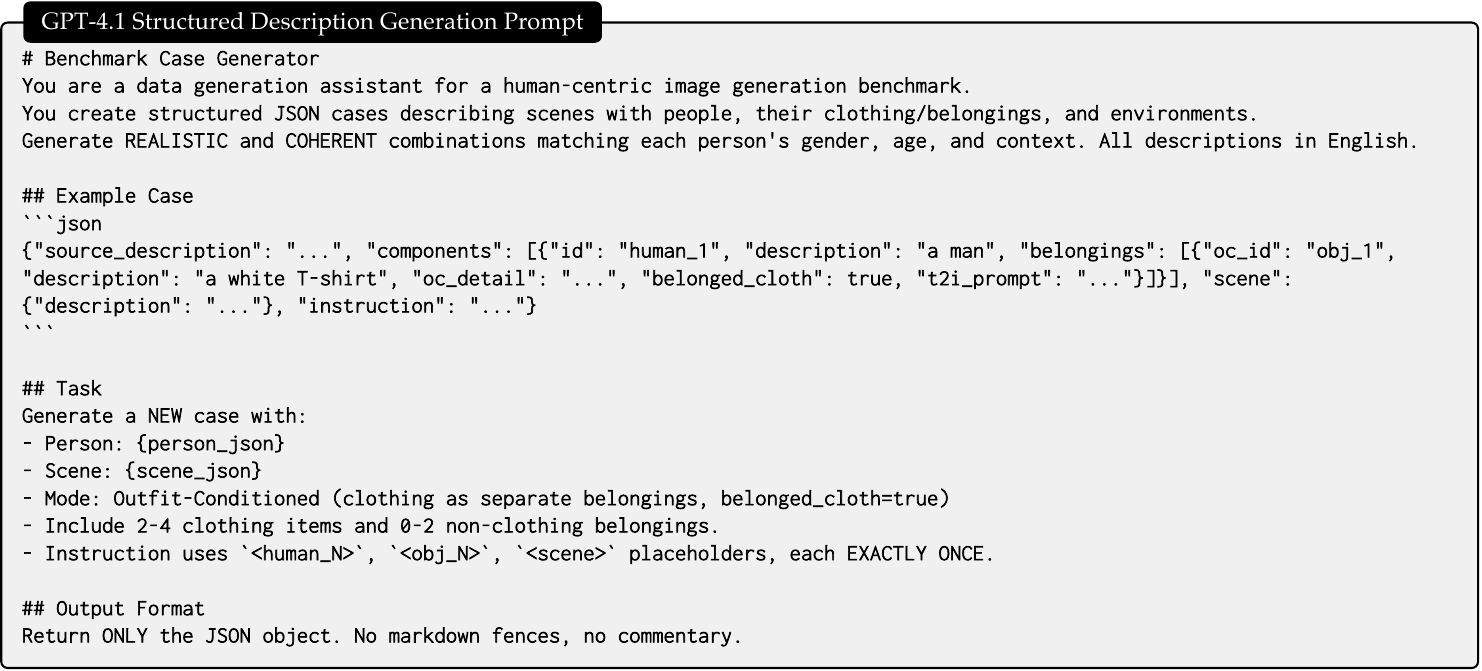}\\
    \vspace{0.2em}
    \includegraphics[width=\textwidth]{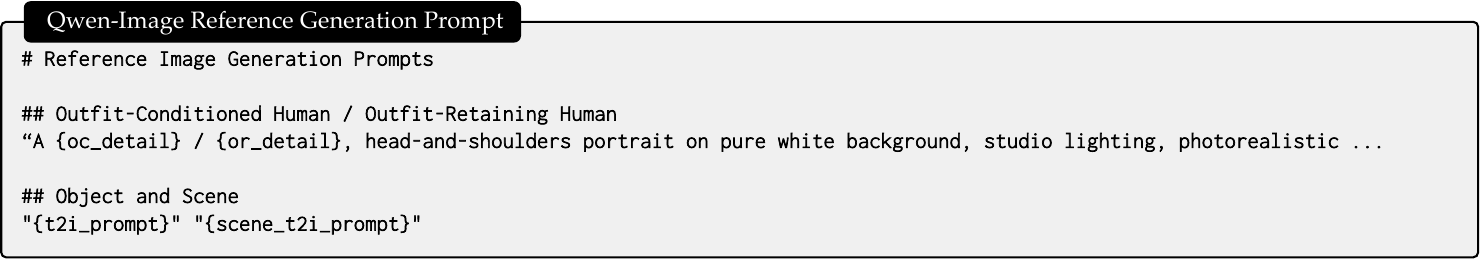}
    % \vspace{-2em}
   \caption{\textbf{(Top)} Scene prototype generation prompt: GPT-4.1~\cite{gpt4-1} produces diverse environment descriptions, each paired with a text-to-image rendering prompt. \textbf{(Middle)} Structured description and instruction synthesis prompt: GPT-4.1~\cite{gpt4-1} synthesizes belonging descriptions, a full-image caption, and both natural-language and identifier-based instructions. \textbf{t2i\_prompt}: text-to-image prompt for reference image generation (see~\cref{fig:step1_vlm_captioning_prompt} for \textbf{oc\_id} and \textbf{oc\_detail}). \textbf{(Bottom)} Reference image generation prompts and examples: head-and-shoulders portraits for Outfit-Conditioned references, three-quarter-length portraits for Outfit-Retaining references, flat-lay renderings for objects, and empty environment renderings for scenes~\cite{wu2025qwen}.}
    \label{fig:step2_llm_scene_prototypes_prompt}
\end{figure*}

\myparagraph{Step 1: Human and Scene Prototype Generation.}
The human prototypes are constructed by exhaustively enumerating combinations of two
gender attributes (\ie, male and female), four age groups (\ie, child, young adult,
middle-aged, and senior), and five ethnicity groups (\ie, Asian, European, Middle
Eastern, African, and Other), yielding 40 distinct human prototypes in total.
Each prototype is associated with a full-body appearance description that additionally
covers clothing, along with dedicated text-to-image prompts for generating both
Outfit-Conditioned and Outfit-Retaining reference images.
The scene prototype pool is constructed by prompting GPT-4.1~\cite{gpt4-1} to generate
400 diverse scene descriptions covering a wide range of indoor and outdoor
settings, each paired with a text-to-image prompt for scene rendering.
The scene prototype generation prompt is illustrated
in~\cref{fig:step2_llm_scene_prototypes_prompt}.

\myparagraph{Step 2: Structured Description and Instruction Synthesis.}
For each benchmark case, GPT-4.1~\cite{gpt4-1} is prompted with the sampled human and
scene prototypes to synthesize a complete structured case description, including belonging descriptions with detailed appearance attributes, a coherent
full-image caption, and both natural-language and identifier-based compositional
instructions. The identifier-based instruction follows the same structured format as the
training data, ensuring consistency between training and evaluation.
The detailed prompt is illustrated in~\cref{fig:step2_llm_scene_prototypes_prompt}.

\begin{figure*}[ht]
    \centering
    \includegraphics[width=\textwidth]{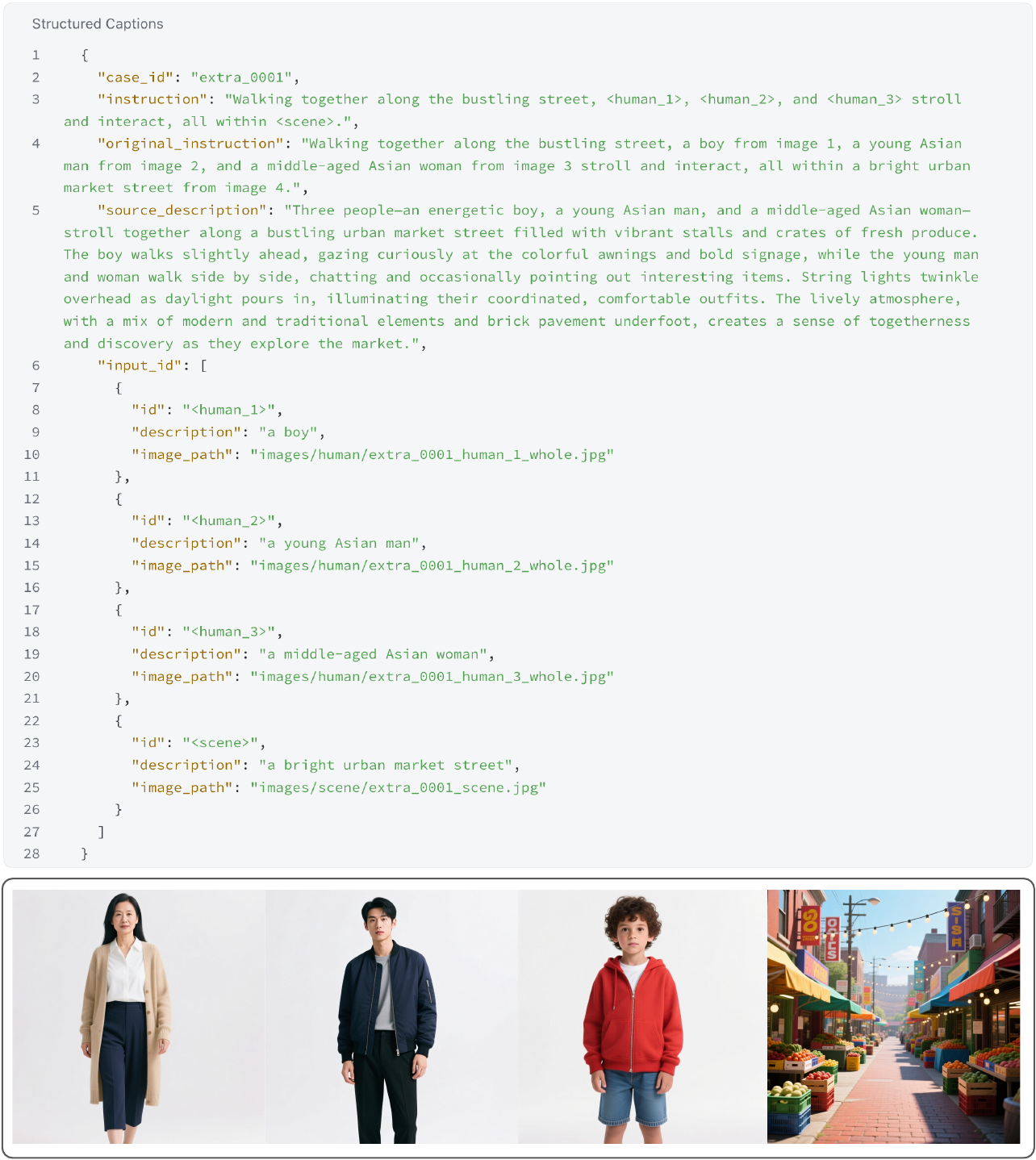}
    \caption{A representative benchmark sample from the multi-person Outfit-Retaining with
    scene subcategory of StructGen Bench. Each sample includes reference images for all
    entities, a natural-language instruction, and an identifier-based instruction.}
    \label{fig:bench_example_1}
\end{figure*}

\begin{figure*}[ht]
    \centering
    \includegraphics[width=\textwidth]{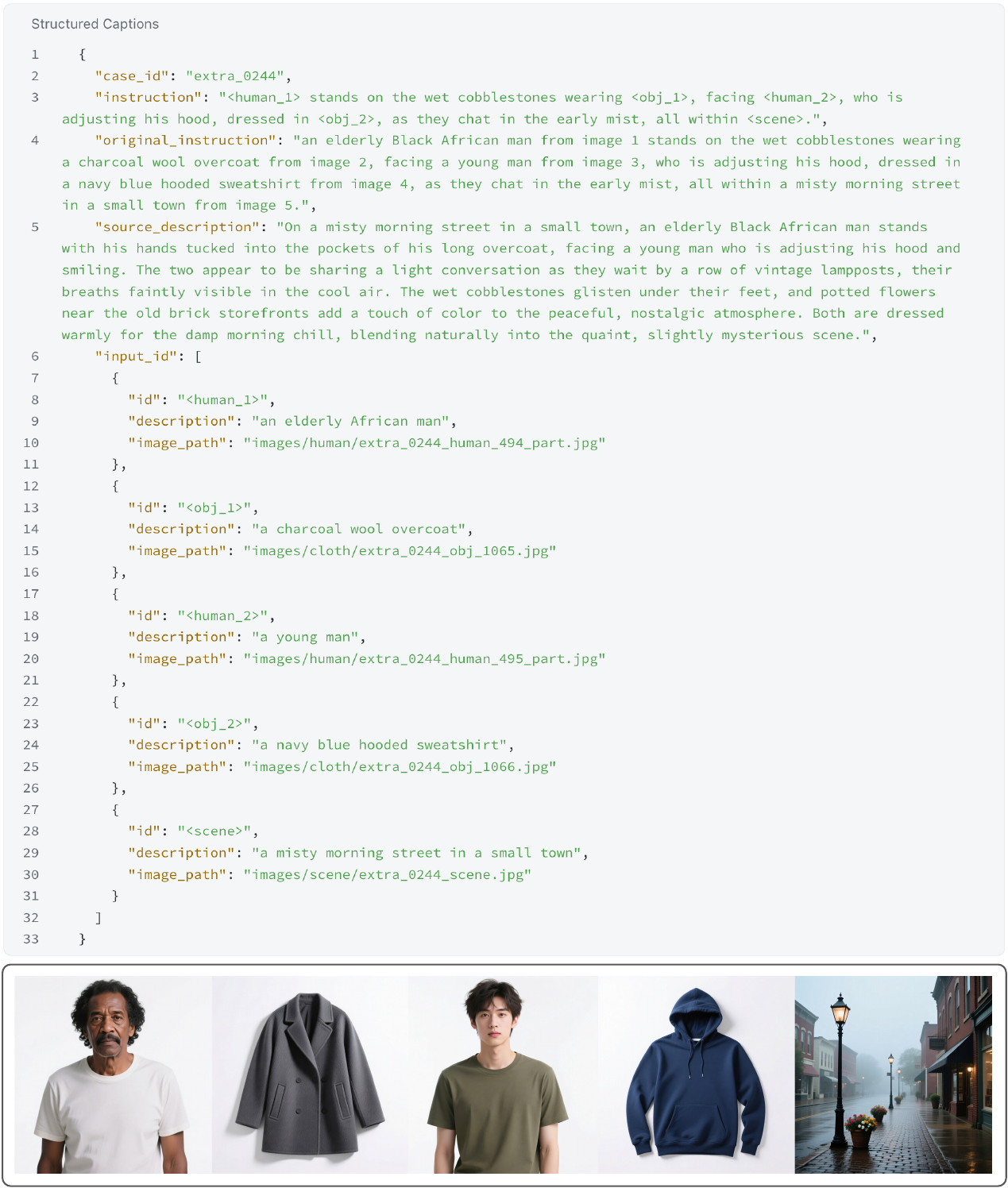}
    \caption{A representative benchmark sample from the multi-person Outfit-Conditioned with
    scene subcategory of StructGen Bench.}
    \label{fig:bench_example_2}
\end{figure*}

\myparagraph{Step 3: Reference Image Generation.}
Reference images for all entities in each benchmark case are generated using
Qwen-Image~\cite{wu2025qwen}, following category-specific templates:
head-and-shoulders portraits for Outfit-Conditioned human references,
three-quarter-length portraits for Outfit-Retaining human references,
flat-lay product renderings for object references, and empty environment renderings
for scene references, as illustrated in~\cref{fig:step2_llm_scene_prototypes_prompt}.
Representative benchmark samples are shown
in~\cref{fig:bench_example_1,fig:bench_example_2}.

\section{Implementation and Training Details}

\myparagraph{Base Model and Architecture.}
StructGen is fine-tuned on top of BAGEL-MICo~\cite{wei2025mico}.
As described in the main paper, the structured context dictionary $\mathcal{D}_{\mathrm{ref}}$
maps each identifier (\eg, $\langle \mathrm{human}_1 \rangle$, $\langle \mathrm{obj}_1 \rangle$,
$\langle \mathrm{scene} \rangle$) to either visual tokens encoded by the ViT and VAE from a
reference image, or textual tokens encoded by the language model from a detailed caption.
This design requires no modification to the base model architecture, enabling flexible mixed-modality reference inputs without any architectural changes.

\myparagraph{Training Configuration.}
Three datasets are mixed and sampled proportionally to their respective sample counts.
For the MICo-De\&Re subset, only the Outfit-Conditioned task variant is used, as this
data does not contain full-body reference images required for the Outfit-Retaining task.
For both our curated subsets, each training sample is randomly instantiated in either
an Outfit-Conditioned or Outfit-Retaining format, enabling the model to jointly learn
fine-grained attribute associating and holistic character composition.
To anchor the latent representation space and prevent catastrophic forgetting, the ViT,
VAE, and MLP connector are frozen throughout training, while the language model backbone,
attention layers, and layer normalization layers remain trainable.
We apply modality-level mixed sampling: each reference entry in
$\mathcal{D}_{\mathrm{ref}}$ is independently sampled between either the image or text
modality with equal probability. When text is sampled, the corresponding reference image
is replaced by its detailed caption from our data pipeline and encoded by the model's
text branch, enabling the model to ground references through both visual and semantic
signals.
The full configuration is summarized below.

\begin{center}
\small
\renewcommand{\arraystretch}{0.85}
\captionof{table}{Training configuration, including datasets, trainable components, sampling strategy, and hyperparameters.}
\label{tab:training_details}
\vspace{2pt}
\begin{tabular}{lp{5cm}}
\toprule
\multicolumn{2}{l}{\textit{Dataset Composition}} \\
\midrule
MICo-De\&Re~\cite{wei2025mico} & 10,665 / Outfit-Conditioned only \\
Single-Person                  & 10,247 / both variants \\
Multi-Person                   &  4,750 / both variants \\
Total                          & 25,662 samples \\
\midrule
\multicolumn{2}{l}{\textit{Frozen / Trainable Components}} \\
\midrule
Frozen    & ViT, VAE, MLP connector \\
Trainable & LLM backbone, attn.\ layers, layer norm \\
\midrule
\multicolumn{2}{l}{\textit{Sampling Strategy}} \\
\midrule
Task-level     & Both variants with equal probability \\
Modality-level & Image / text with equal probability \\
\midrule
\multicolumn{2}{l}{\textit{Hyperparameters}} \\
\midrule
GPU            & 4\,$\times$\,H20 (FSDP FULL\_SHARD) \\
Train steps    & 13{,}000 \\
Learning rate  & $2 \times 10^{-5}$ \\
LR schedule    & Constant \\
Warmup steps   & 200 \\
Weight decay   & 0 \\
Batch tokens   & 45{,}000 \\
Max tokens     & 45{,}000 \\
Precision      & bf16 \\
\bottomrule
\end{tabular}
\end{center}

\section{Qualitative Comparisons with Baseline Model}

\begin{figure*}[ht]
    \centering
    \includegraphics[width=\textwidth]{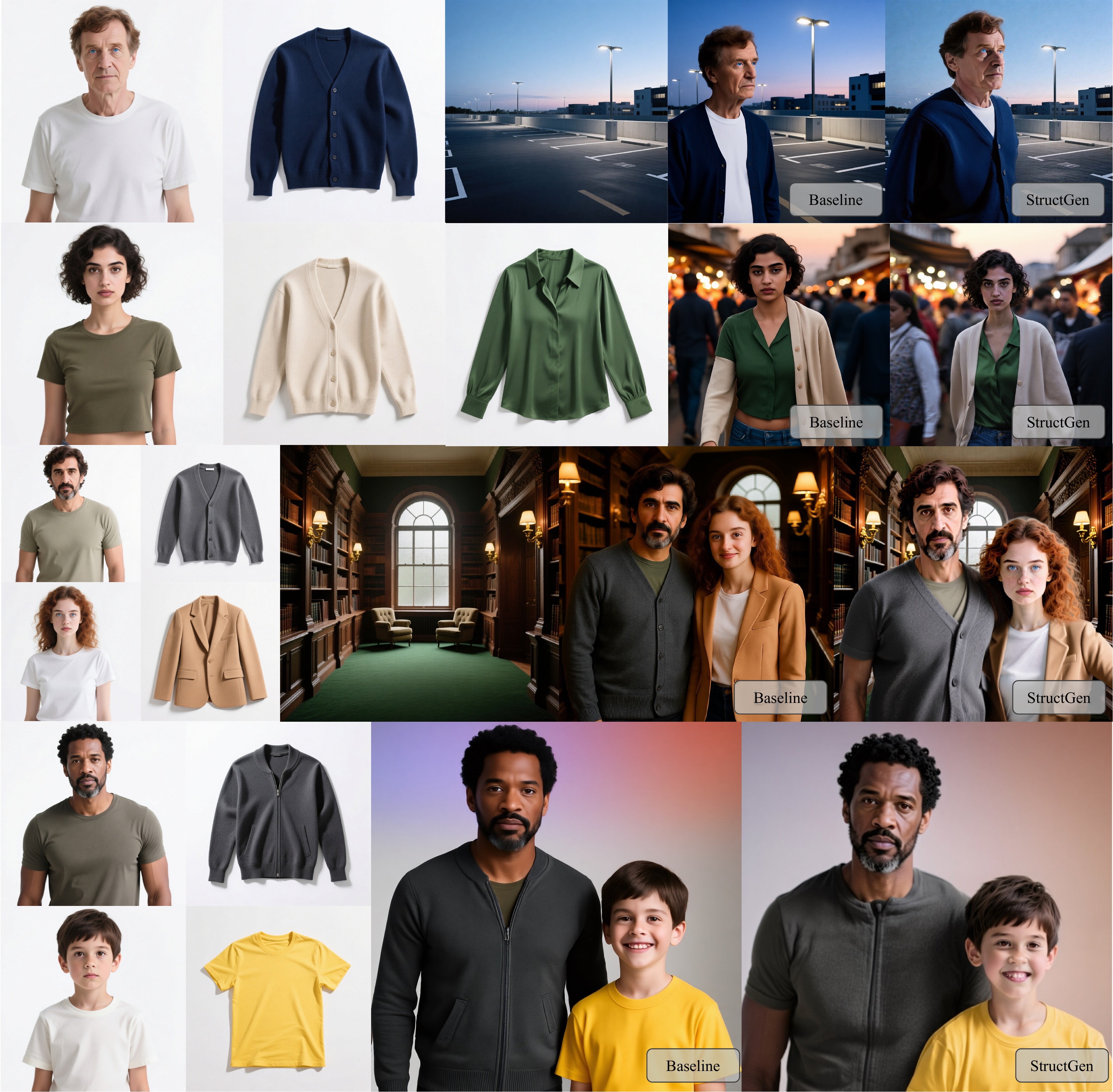}
    \caption{Qualitative comparisons with Echo-4o~\cite{ye2025echo} on Outfit-Conditioned cases,
    where the model must compose a target image from separate human identity and clothing
    references. StructGen achieves more accurate attribute--subject association and stronger
    consistency with all references.}
    \label{fig:comp_outfit_conditioned}
\end{figure*}

\begin{figure*}[ht]
    \centering
    \includegraphics[width=\textwidth]{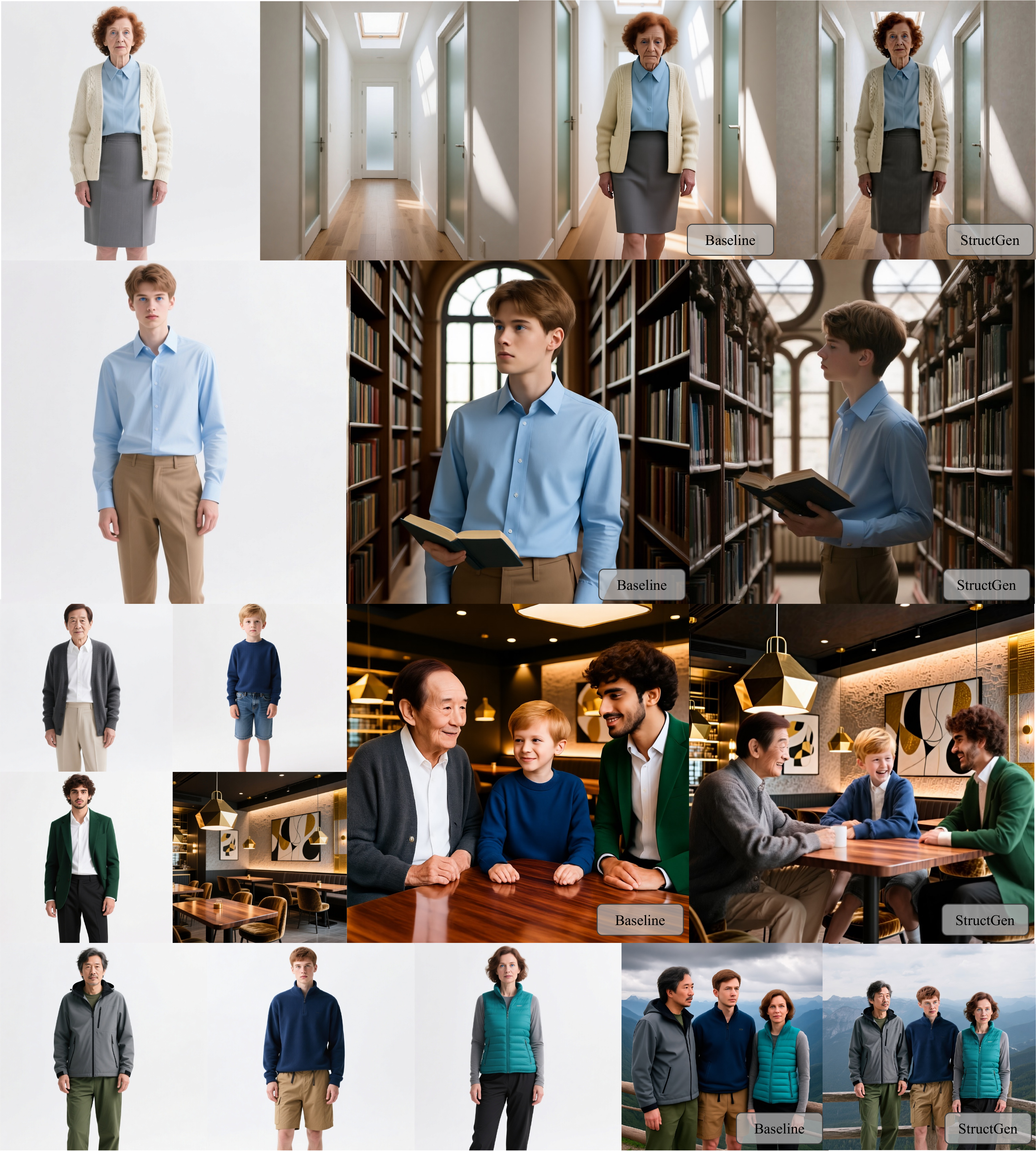}
    \caption{Qualitative comparisons with Echo-4o~\cite{ye2025echo} on Outfit-Retaining cases,
    where the model must faithfully preserve the full appearance of referenced human subjects
    including their original outfits. StructGen better preserves both facial identity and
    full-body appearance consistency.}
    \label{fig:comp_outfit_retaining}
\end{figure*}

We provide additional qualitative comparisons between StructGen and Echo-4o~\cite{ye2025echo}.
Both models are evaluated on the same reference images and generation instructions from
StructGen Bench, using the natural-language instruction for Echo-4o and the
identifier-based for StructGen.
\cref{fig:comp_outfit_conditioned} presents Outfit-Conditioned cases and
\cref{fig:comp_outfit_retaining} presents Outfit-Retaining cases.

\end{multicols}
% -------- Section 1: Data Curation Pipeline --------

% -------- Section 2: Benchmark Pipeline --------

% -------- Section 4: Qualitative Comparisons --------

\end{document}